\documentclass[11pt]{article}

\pdfoutput=1
\usepackage{geometry}
\usepackage{microtype}
\usepackage{graphicx}
\usepackage{multirow}
\usepackage{subfigure}
\usepackage{booktabs} 

\usepackage{hyperref}

\usepackage{algorithm}
\usepackage{algorithmicx,algpseudocode}

\makeatletter
\def\ALG@special@indent{%
    \ifdim\ALG@thistlm=0pt\relax
        \hskip-\leftmargin
    \else
        \hskip\ALG@thistlm
    \fi
}
\algnewcommand\algorithmicinput{\textbf{Input:}}
\algnewcommand\Input{\item[\algorithmicinput]}

\renewcommand{\algorithmicrequire}{\textbf{Input:}} 

\usepackage[utf8]{inputenc} 
\usepackage[T1]{fontenc}    
\usepackage{url}
\usepackage{pifont}
\usepackage{amsmath}
\usepackage{subfigure}
\usepackage{caption}
\usepackage{hyperref} 

\def\plambda{{\lambda}_{\perp}}
\usepackage{relsize}
\def\btheta{\hat \theta}

\usepackage{mathtools}
\usepackage{xcolor,colortbl}
\def\E{{\mathbb E}}
\usepackage{amssymb}

\newenvironment{itemize*}%
{\begin{itemize}[leftmargin=*,topsep=0pt]%
		\setlength{\itemsep}{1pt}%
		\setlength{\parskip}{1pt}}%
	{\end{itemize}}

\usepackage{enumerate}
\usepackage{enumitem}
\def\R{{\mathbb{R}}}
\usepackage{mathtools}
\newcommand{\citep}[1]{\cite{#1}}
\newcommand{\norm}[1]{\left\lVert#1\right\rVert}
\newcommand{\nunorm}[1]{\left\lVert#1\right\rVert_{\text{nuc}}}
\newcommand{\opnorm}[1]{\left\lVert#1\right\rVert_{\text{op}}}
\newcommand{\vecc}[1]{\text{vec}(#1)}
\newcommand{\inp}[2]{\langle #1,#2 \rangle}
\newcommand{\new}{\text{new}}

\algnewcommand{\IfThenElse}[3]{ \IfThenElse{<if>}{<then>}{<else>}
  \STATE \algorithmicif\ #1\ \algorithmicthen\ #2\ \algorithmicelse\ #3}

\newcommand{\Stage}[1]{\item[]\noindent\ALG@special@indent \textbf{Stage}\ #1}
\usepackage{amsthm}

\usepackage[normalem]{ulem}
\usepackage[export]{adjustbox}
\theoremstyle{plain}
\newtheorem{theorem}{Theorem}[section]
\newtheorem{proposition}[theorem]{Proposition}
\newtheorem{lemma}[theorem]{Lemma}
\newtheorem{corollary}[theorem]{Corollary}
\theoremstyle{definition}
\newtheorem{definition}[theorem]{Definition}
\newtheorem{assumption}[theorem]{Assumption}
\theoremstyle{remark}
\newtheorem{remark}[theorem]{Remark}

\usepackage[textsize=tiny]{todonotes}
\usepackage{environ}
\newcommand{\acksection}{\section*{Acknowledgments and Disclosure of Funding}}
\NewEnviron{ack}{%
  \acksection
  \BODY
}

\title{Efficient Frameworks for Generalized Low-Rank Matrix Bandit Problems}

\author
{
Yue Kang\thanks{Department of Statistics, University of California, Davis; e-mail: {\tt yuekang@ucdavis.edu}} 
	\and 
 Cho-Jui Hsieh\thanks{Department of Computer Science, University of California, Los Angeles; e-mail: {\tt chohsieh@cs.ucla.edu}} 
	\and
 Thomas C. M. Lee\thanks{Department of Statistics, University of California, Davis; e-mail: {\tt tcmlee@ucdavis.edu}} 
}

\begin{document}

\date{}
\maketitle


\begin{abstract}
In the stochastic contextual low-rank matrix bandit problem, the expected reward of an action is given by the inner product between the action's feature matrix and some fixed, but initially unknown $d_1$ by $d_2$ matrix $\Theta^*$ with rank $r \ll \{d_1, d_2\}$, and an agent sequentially takes actions based on past experience to maximize the cumulative reward. In this paper, we study the generalized low-rank matrix bandit problem, which has been recently proposed in \cite{lu2021low} under the Generalized Linear Model (GLM) framework. To overcome the computational infeasibility and theoretical restrain of existing algorithms on this problem, we first propose the G-ESTT framework that modifies the idea from \cite{jun2019bilinear} by using Stein's method on the subspace estimation and then leverage the estimated subspaces via a regularization idea. Furthermore, we remarkably improve the efficiency of G-ESTT by using a novel exclusion idea on the estimated subspace instead, and propose the G-ESTS framework. We also show that G-ESTT can achieve the $\tilde{O}(\sqrt{(d_1+d_2)MrT})$ bound of regret while G-ESTS can achineve the $\tilde{O}(\sqrt{(d_1+d_2)^{3/2}Mr^{3/2}T})$ bound of regret under mild assumption up to logarithm terms, where $M$ is some problem dependent value. Under a reasonable assumption that $M = O((d_1+d_2)^2)$ in our problem setting, the regret of G-ESTT is consistent with the current best regret of $\tilde{O}((d_1+d_2)^{3/2} \sqrt{rT}/D_{rr})$~\citep{lu2021low} ($D_{rr}$ will be defined later). For completeness, we conduct experiments to illustrate that our proposed algorithms, especially G-ESTS, are also computationally tractable and consistently outperform other state-of-the-art (generalized) linear matrix bandit methods based on a suite of simulations.
\end{abstract}
\section{Introduction}\label{sec:intro}

The contextual bandit has proven to be a powerful framework for sequential decision-making problems, with great applications to clinical trials~\citep{woodroofe1979one}, recommendation system~\citep{li2010contextual}, and personalized medicine~\citep{bastani2020online}. This class of problems evaluates how an agent should choose an action
from the potential action set at each round based on an updating policy on-the-fly so as to maximize the cumulative reward or minimize the overall regret. 
With high dimensional sparse data becoming ubiquitous in various fields nowadays, the most fundamental (generalized) linear bandit framework, although has been extensively studied, becomes inefficient in practice. This fact consequently leads to a line of work on stochastic high dimensional bandit problems with low dimensional structures~\citep{johnson2016structured,li2021simple}, such as the LASSO bandit and low-rank matrix bandit.
\par In this work, we investigate on the generalized low-rank matrix bandit problem firstly studied in~\cite{lu2021low}: at round $t = 1,\dots,T$, the algorithm selects an action represented by a $d_1$ by $d_2$ matrix $X_t$ from the admissible action set $\mathcal{X}_t$ ($\mathcal{X}_t$ may be fixed), and receives its associated noisy reward $y_t = \mu(\inp{\Theta^*}{X_t}) + \eta_t$ where $\Theta^* \in \mathbb{R}^{d_1 \times d_2}$ is some unknown low-rank matrix with rank $r \ll \{d_1,d_2\}$ and $\mu(\cdot)$ is the inverse link function. More details about 
this setting are deferred to Section \ref{sec:pre}. 
This problem has vast applicability in real world applications. On the one hand, matrix inputs are appropriate when dealing with paired contexts which are omnipresent in practice. For instance,
to design a personalized movie recommendation system, 
we can formulate each user as $m$ $d_1$-dimensional feature vectors ($x_1,\dots,x_m \in \mathbb{R}^{d_1}$) and each movie as $m$ $d_2$-dimensional feature vectors ($y_1,\dots,y_m \in \mathbb{R}^{d_2}$). A user-item pair can then be naturally represented by a feature matrix defined as the summation of the outer products $\sum_{k=1}^m x_ky_k^\top \in \mathbb{R}^{d_1 \times d_2}$, which will become the contextual feature observed by the bandit algorithm.
Other applications involve interaction features between two groups, such as flight-hotel bundles~\citep{lu2021low} and dating service~\citep{jun2019bilinear} can  also be similarly established. Besides, low-rank models have gained tremendous success in various areas~\citep{candes2009exact}. In particular, our problem can be regarded as an extension of the  inductive matrix factorization problem~\citep{jain2013provable,zhong2015efficient}, which estimates low-rank matrices with contextual information, under the online learning scenario. 
\par Our study is inspired by a line of work on stochastic contextual low-rank matrix bandit~\citep{jang2021improved,jun2019bilinear,lu2021low}.
To design an algorithm for matrix bandit problems, a na\"ive approach is to flatten the $d_1$ by $d_2$ feature matrices into vectors and then apply any (generalized) linear bandit algorithms, which, however, would be inefficient when $d_1d_2$ is large. To take advantage of the low-rank structure, \cite{jun2019bilinear} have introduced the bilinear low-rank bandit problem and proposed a two-stage algorithm named ESTR which could achieve a regret bound of $\tilde O((d_1+d_2)^{3/2}\sqrt{rT}/D_{rr})$\footnote{$\tilde{O}$ ignores the polylogarithmic factors.}. 
Subsequently,~\cite{jang2021improved} constructed a new algorithm called $\epsilon$-FALB for bilinear bandits and achieved a better regret of $\tilde{O}(\sqrt{d_1d_2(d_1+d_2)T})$. However, they only studied the linear reward framework and
also restricted the feature matrix as a rank-one matrix. As a follow-up work, \cite{lu2021low} further released the rank-one restriction on the action feature matrices, and they introduced an algorithm LowGLOC based on the online-to-confidence-set conversion~\citep{abbasi2012online} for generalized low-rank matrix bandits with $\tilde{O}(\sqrt{(d_1+d_2)^3rT})$ regret bound. However, this method can't handle the contextual setting since the arm set is assumed fixed at each round. This algorithm is also computationally prohibitive since it requires to calculate the weights of a self-constructed covering of the admissible parameter space at each iteration. And how to find this covering for low-rank matrices is also unclear. 

In this work, we propose two efficient methods called G-ESTT and G-ESTS for this problem by modifying two stages of ESTR appropriately from different perspectives. To the best of our knowledge, the proposed methods are the first two generalized (contextual) low-rank bandit algorithms that are computationally feasible, and achieve the decent regret bound of $\tilde{O}(\sqrt{(d_1+d_2)^3 rT}/D_{rr})$ and $\tilde{O}({(d_1+d_2)^{7/4} r^{3/4}T}/D_{rr})$ on low-rank bandits. 
The main contributions of this paper can be summarized as: \textbf{1)} we propose two novel two-stage frameworks G-ESTT and G-ESTS
under some mild assumptions. Compared with ESTR in~\cite{jun2019bilinear}, $\epsilon$-FALB in~\cite{jang2021improved} and LowESTR in~\cite{lu2021low}, our algorithms are proposed for the nonlinear reward framework with arbitrary action matrices. Compared with LowGLOC in~\cite{lu2021low}, our algorithms are computationally feasible in practice. \textbf{2)} For G-ESTT, we extend the GLM-UCB algorithms~\citep{filippi2010parametric} via a novel regularization technique. 
\textbf{3)} Our proposed G-ESTS is simple and could be easily implemented based on any state-of-the-art misspecified linear bandit algorithms to achieve the regret bound of order $\tilde{O}{(d_1+d_2)^{7/4} r^{3/4}T}/D_{rr})$. In practice, it can be used with any generalized linear bandit to achieve high efficiency. Particularly, when we combine G-ESTS with some efficient algorithms (e.g. SGD-TS~\citep{ding2021efficient}), the total time complexity after a warm-up stage scales as $O(Tr(d_1+d_2))$. \textbf{4)} The practical superiority of our algorithms are firmly validated based on my experimental results.
\newline \textbf{Notations:} For a vector $x \in \mathbb{R}^n$, we use $\norm{x}_p$ to denote the $l_p$-norm of the vector $x$ and $\norm{x}_H = \sqrt{x^\top H x}$ to denote its weighted $2-$norm with regard to a positive definite matrix $H \in \mathbb{R}^{n \times n}$. For matrices $X, Y \in \mathbb{R}^{n_1 \times n_2}$, we use $\norm{X}_{\text{op}}$, $\nunorm{X}$ and $\norm{X}_F$ to define the operator norm, nuclear norm and Frobenious norm of matrix $X$ respectively, and we denote $\inp{X}{Y} \coloneqq \textbf{trace}(X^\top Y)$ as the inner product between $X$ and $Y$. We write $f(n) \asymp g(n)$ if $f(n) = O(g(n))$ and $g(n) = O(f(n))$.


\section{Related Work}\label{sec:relatedwork}

In this section, we briefly discuss some previous algorithms on low-rank matrix bandit problems. Besides the works we have discussed in the former section, \cite{katariya2017bernoulli,trinh2020solving} considered the rank-one bandit problems where the expected reward forms a rank-one matrix and the player selects an element from this matrix as the expected reward at each round. In addition, \cite{katariya2017stochastic} also studied the rank-one matrix bandit via an elimination-based algorithm. Alternatively, \cite{gopalan2016low,kveton2017stochastic,lu2018efficient} considered the general low-rank matrix bandit, and furthermore \cite{hao2020low} considered a stochastic low-rank tensor bandit. However, for all these works the feature matrix of an action could be flattened into a one-hot basis vector, and our work yields a more general structure.
\par Additionally, \cite{li2021simple} extended some previous works~\citep{johnson2016structured} and presented a unified algorithm based on a greedy search for high-dimensional bandit problems. But it's nontrivial to extend the framework to the matrix bandit problem. For example, they assume that the minimum eigenvalue of the covariance matrix 
could be strictly lower bounded, but this lower bound would mostly depend on the size of feature matrices, and hence would affect the regret bound consequently. 

\section{Preliminaries}\label{sec:pre}

In this section we review our problem setting and introduce the assumptions for our theoretical analysis. 
Let $T$ be the total number of rounds and $\mathcal{X}_t$ be the action set ($\mathcal{X}_t$ could be fixed or not). Throughout this paper,  we denote the action set $\mathcal{X}_t = \mathcal{X}$ as fixed for notation simplicity, while our frameworks also work with the same regret bound when $\mathcal{X}_t$ varies over time (see Appendix \ref{context} for more details.) Algorithms along with theory could be identically obtained when the action set varies (Appendix \ref{context}). 
At each round $t \in [T]$, The agent selects an action $X_{t} \in \mathcal{X}_t$ and gets the payoff $y_t$ which is conditionally independent of the past payoffs and choices. For the generalized low-rank matrix bandits, we assume the payoff $y_t$ follows a canonical exponential family such that:
\begin{gather}
    p_{\Theta^*} (y_t | X_t) = \exp\left({\frac{y_t \beta -b(\beta)}{\phi}+ c(y_t,\phi)} \right), \text{  where } \ \beta = \vecc{X_t}^\top \vecc{\Theta^*} \coloneqq \inp{X_t}{\Theta^*}, \label{density}  \\
    \E_{\Theta^*} (y_t | X_t) = b^{\prime} (\inp{X_t}{\Theta^*}) \coloneqq \mu (\inp{X_t}{\Theta^*}), \nonumber 
\end{gather} 
where $\Theta^* \subseteq \mathit{\Theta}$ is a fixed but unknown matrix with rank $r \ll \{d_1,d_2\}$ and $\mathit{\Theta}$ is some admissible compact subset of $\mathbb{R}^{d_1 \times d_2}$ (w.l.o.g. $d_1 = \Theta(d_2)$). We also call $\mu (\inp{X_t}{\Theta^*})$ the reward of action $X_t$.

In addition, one can represent model \eqref{density} in the following Eqn. \eqref{subgform}.
Note that if we relax the definition of $\mu(\cdot)$ to any real univariate function with some centered exogenous random noise $\eta_t$, the model shown in Eqn. \eqref{subgform} generalizes our problem setting to a single index model (SIM) matrix bandit, and the generalized low-rank matrix bandit problem is a special case of this model.
\begin{align}
    y_t = \mu (\inp{X_t}{\Theta^*}) + \eta_t. \label{subgform}
\end{align}
Here, $\eta_t$ follows the sub-Gaussian property with some constant parameter $\sigma_0$ conditional on the filtration $\mathcal{F}_t = \{X_t,X_{t-1},\eta_{t-1},\dots,X_1,\eta_1\}$. We also denote $d = \max\{d_1,d_2\}$. And it is natural to evaluate the agent's strategy based on the regret~\citep{audibert2009exploration}, defined as the difference between the total reward of optimal policy and the agent's total reward in practice:
\begin{align*}
    Regret_{t} = \sum_{i=1}^t \max_{X \in \mathcal{X}}\mu(\inp{X}{\Theta^*})-\mu(\inp{X_i}{\Theta^*}).
\end{align*}
We also present the following two definitions to facilitate further analysis via Stein's method:
\begin{definition}\label{def:score}
Let $p:\mathbb{R} \rightarrow \mathbb{R}$ be a univariate probability density function defined on $\mathbb{R}$. The score function $S^p:\mathbb{R} \rightarrow \mathbb{R}$ regarding density $p(\cdot)$ is defined as:
$$
    S^p(x) = -\nabla_{x} log(p(x)) = -\nabla_x p(x)/p(x), \quad \; x \in \R.
$$
In particular, for a random matrix with its entrywise probability density $\mathbf{p} = (p_{ij}):\mathbb{R}^{d_1\times d_2} \rightarrow \mathbb{R}^{d_1\times d_2}$, we define its score function $S^\mathbf{p} = (S^\mathbf{p}_{ij}):\mathbb{R}^{d_1\times d_2} \rightarrow \mathbb{R}^{d_1\times d_2}$ as $S^\mathbf{p}_{ij}(x) = S^{p_{ij}}(x)$ by applying the univariate score function to each entry of $\mathbf{p}$ independently.
\end{definition}
\begin{definition}\label{def:dilation}\textup{(Fact 2.6,~\citep{minsker2018sub})}
Given a rectangular matrix $A \in \mathbb{R}^{d_1\times d_2}$, the (Hermitian) dilation $\mathcal{H}:\mathbb{R}^{d_1\times d_2} \rightarrow \mathbb{R}^{(d_1+d_2)\times (d_1+d_2)}$ is defined as:
$$\mathcal{H}(A) = \begin{pmatrix}
0 & A\\
A^\top & 0 
\end{pmatrix}.$$
\end{definition}
We would omit the subscript $x$ of $\nabla$ and the superscript $p$ of $S$ when the underlying distribution is clear. With these definitions, we make the following mild assumptions:
\begin{assumption} \textup{(Finite second-moment score)} \label{assu_sampling}
There exists a sampling distribution $\mathcal{D}$ over $\mathcal{X}$ such that for the random matrix $X$ drawn from $\mathcal{D}$ with its associated density $\mathbf{p}:\mathbb{R}^{d_1\times d_2} \rightarrow \mathbb{R}^{d_1\times d_2}$, we have $\mathbb{E}[(S^{\mathbf{p}}(X))_{ij}^2] \leq M, \forall i,j$. And the columns or rows of  random matrix $X$ are pairwisely independent.
\end{assumption}

\begin{assumption} \label{assu_bound}
The norm of true parameter $\Theta^*$ and feature matrices in $\mathcal{X}$ is bounded: there exists $S \in \mathbb{R}^+$ such that for all arms $X \in \mathcal{X}$, $\norm{X}_F, \norm{\Theta^*}_F \leq S_0$. 
\end{assumption}
\begin{assumption} \label{assu_link}
The inverse link function $\mu(\cdot)$ in GLM is continuously differentiable and there exist two constants $c_\mu,k_\mu$ such that $0 < c_\mu \leq \mu^\prime (x) \leq k_\mu$ for all $|x| \leq S_0$.
\vspace{-0.1 cm}
\end{assumption}

Assumption \ref{assu_sampling} is commonly used in Stein's method~\citep{chen2010normal}, and easily satisfied by a wide range of distributions that are non-zero-mean or even non sub-Gaussian thereby allowing us to work with cases not previously possible. For example, to find $\mathcal{D}$ we only need the convex hull of $\mathcal{X}$ contains a ball with radius $R$, and then we can use $p_{ij}$ as centered normal p.d.f. with variance $R^2/(d_id_j)$. This choice works well in our experiments and please refer to Appendix \ref{sec:app:exp} for more details.
Furthermore, Assumption \ref{assu_bound} and \ref{assu_link} are also standard in contextual generalized bandit literature, and they explicitly imply that we have an upper bound as $|\mu(\inp{X}{\Theta})| \leq |\mu(0)|+k_\mu S_0 \coloneqq S_f$.



\section{Main Results}
In this section, we present our novel two-stage frameworks, named Generalized Explore Subspace Then Transform (G-ESTT) and Generalized Explore Subspace Then Subtract (G-ESTS) respectively. These two algorithms are inspired by the two-stage algorithm ESTR proposed in~\cite{jun2019bilinear}. ESTR estimates the row and column subspaces for the true parameter $\Theta^*$ in stage 1. In stage 2, it exploits the estimated subspaces and transforms the original matrix bandits into linear bandits with 
sparsity, and then invoke a penalized approach called LowOFUL.

To extend their work into the nonlinear reward setting, we firstly adapt stage 1 into the GLM framework by estimating $\Theta^*$ via the following quadratic optimization problem in Eqn. \eqref{loss} inspired by a line of work in nonlinear signal estimation~\citep{plan2016generalized,yang2017high}. And then we propose two different methods based on the estimated subspaces to deal with GLM bandits for stage 2:
for G-ESTT we adapt the idea of LowOFUL into the GLM framework and propose an improved algorithm with regularization. For G-ESTS we innovatively drop all negligible entries to get a low-dimensional bandit, which could efficiently be solved by any modern generalized linear bandit algorithm with much less computation. 
\subsection{Stage 1: Subspace Exploration}\label{subsec:stage1}
\begin{algorithm}[t]
\caption{Generalized Explore Subspace Then Transform (G-ESTT)} \label{alg_estt}
\begin{algorithmic}[1]
\Input $\mathcal{X}, T, T_1, \mathcal{D}$, the probability rate $\delta$, parameters for Stage 2: $\lambda, \lambda_\perp$. \vspace{0.07 cm}
\Stage \textbf{1: Subspace Estimation}
\For{$t=1$ {\bfseries to} $T_1$}
\State Pull arm $X_t \in \mathcal{X}$ according to $\mathcal{D}$, observe payoff $y_t$.
\EndFor
\State Obtain $\widehat \Theta$ based on Eqn. \eqref{loss}.
\State Obtain the full SVD of $\widehat \Theta = [\widehat U, \widehat U_\perp] \, \widehat D \, [\widehat V, \widehat V_\perp]^\top$ where $\widehat U \in \mathbb{R}^{d_1 \times r}, \widehat V \in \mathbb{R}^{d_2 \times r}$. \vspace{0.07 cm}
\Stage \textbf{2: Sparse Generalized Linear Bandits}
\State Rotate the arm feature set: $\mathcal{X}^\prime \coloneqq [\widehat U, \widehat U_\perp]^\top \mathcal{X} [\widehat V, \widehat V_\perp]$ and the admissible parameter space: $\mathit{\Theta}^{\prime} \coloneqq [\widehat U, \widehat U_\perp]^\top \mathit{\Theta} [\widehat V, \widehat V_\perp]$.
\State Define the vectorized arm set so that the last $(d_1-r) \cdot (d_2-r)$ components are negligible:
\begin{align}
    \mathcal{X}_0 \coloneqq \{\vecc{\mathcal{X}^\prime_{1:r,1:r}},\vecc{\mathcal{X}^\prime_{r+1:d_1,1:r}},  \vecc{\mathcal{X}^\prime_{1:r,r+1:d_2}},\vecc{\mathcal{X}^\prime_{r+1:d_1,r+1:d_2}}\}, \label{rotate_arm}
\end{align}
and similarly define the parameter set:
\begin{align}
    \mathit{\Theta}_0 \coloneqq \{\vecc{\mathit{\Theta}^\prime_{1:r,1:r}},\vecc{\mathit{\Theta}^\prime_{r+1:d_1,1:r}}, \vecc{\mathit{\Theta}^\prime_{1:r,r+1:d_2}},\vecc{\mathit{\Theta}^\prime_{r+1:d_1,r+1:d_2}}\}. \label{rotate_theta}
\end{align}
\State For $T_2 = T -T_1$ rounds, invoke (P)LowGLM-UCB with $\mathcal{X}_0,\mathit{\Theta}_0,k = (d_1+d_2)r-r^2,(\lambda_0,\lambda_\perp)$.
\end{algorithmic}
\end{algorithm}
\vspace{-3 mm}
For any real-value function $f(\cdot)$ defined on $\R$, and symmetric matrix $A \in \R^{d \times d}$ with its SVD decomposition as $A=UDU^\top$, we define $f(A) \coloneqq U \, \text{diag}(f(D_{11}),\dots,f(D_{dd})) \, U^\top$.
To explore the valid subspace of the parameter matrix $\Theta^*$, we firstly define a function $\psi:\mathbb{R} \rightarrow \mathbb{R}$~\citep{minsker2018sub} in Eqn. \eqref{psi} and subsequently we define $\tilde\psi_\nu: \mathbb{R}^{d_1 \times d_2} \rightarrow \mathbb{R}^{d_1 \times d_2}$ as $\tilde\psi_\nu(A) = \psi(\nu \mathcal{H}(A))_{1:d_1, (d_1+1):(d_1+d_2)} / \nu$ for some parameter $\nu \in \mathbb{R}^+$.
\begin{align}\label{psi}
\psi(x) = 
\begin{cases}
\log(1+x+x^2/2), & \quad x\geq 0; \\
-\log(1-x+x^2/2), & \quad x<0.
\end{cases}
\end{align}
We consider the following well-defined regularized minimization problem with nuclear norm penalty:
\begin{align}
    \widehat \Theta = \arg \min_{\Theta \in \mathbb{R}^{d_1 \times d_2}} L_{T_1}(\Theta) 
    + \lambda_{T_1} \nunorm{\Theta}, \; \; L_{T_1}(\Theta) = \inp{\Theta}{\Theta} - \frac{2}{T_1}\sum_{i=1}^{T_1} \inp{\tilde\psi_\nu(y_i \cdot  S(X_i))}{\Theta}. \label{loss}
\end{align}
An interesting fact is that our estimator is invariant under different choices of function $\mu(\cdot)$, and we could present the following oracle inequality regarding the estimation error $\norm{\widehat\Theta - \mu^*\Theta^*}_F$ for some nonzero constant $\mu^*$ by adapting generalized Stein's Method \cite{chen2010normal}.

\begin{theorem} \label{thm_rscbound} \textup{(Bounds for GLM)}  For any low-rank generalized linear model with samples $X_1\dots,X_{T_1}$ drawn from $\mathcal{X}$ according to $\mathcal{D}$ in Assumption \ref{assu_sampling}, and assume Assumption \ref{assu_bound} and \ref{assu_link} hold,  then for the optimal solution to the nuclear norm regularization problem \eqref{loss} with $\nu = \sqrt{2\log(2(d_1+d_2)/\delta)/((4\sigma_0^2 + S_f^2)MT_1(d_1+d_2))}$ and  $$\lambda_{T_1} = 4\sqrt{\frac{2(4\sigma_0^2+S_f^2)M(d_1+d_2)\log(2(d_1+d_2)/\delta)}{T_1}},$$
with probability at least $1-\delta$ it holds that:
\begin{align}
    &\norm{\widehat \Theta - \mu^*\Theta^*}_F^2 \leq \dfrac{C_1M(d_1+d_2)r\log(\frac{2(d_1+d_2)}{\delta})}{T_1}, \label{thmeq_bound} 
\end{align}
for $C_1=36(4\sigma_0^2+S_f^2)$ and some nonzero constant $\mu^*$.
\end{theorem}


The proof of Theorem \ref{thm_rscbound} is based on a novel adaptation of Stein-typed Lemmas and is deferred to Appendix \ref{app:stein}. 
We believe this oracle bound is non-trivial since the rate of convergence is no worse than that deduced from the restricted strong convexity (details in Appendix \ref{sec:app:rsc}) given $M = O((d_1+d_2)^2)$, even without the regular sub-Gaussian assumption. For completeness, we also present detailed proof of the matrix recovery rate with the restricted strong convexity in Appendix \ref{sec:app:rsc} which may be of separate interest. And its proof is highly different from the one used in the simple linear case~\cite{lu2021low}. We also present an intuitive explanation on why our Stein-type method works well under our problem setting in Appendix \ref{app:rsc:thm} even without the sub-Gaussian assumption. In addition, this bound also holds under a more general SIM in Eqn. \eqref{subgform} other than just GLM. Furthermore, although there exists a non-zero constant $\mu^*$ in the error term, it will not affect the singular vectors and subspace estimation of $\Theta^*$ at all.


After acquiring the estimated $\widehat \Theta$ in stage 1, we can obtain the corresponding SVD as $\widehat \Theta = [\widehat U, \widehat U_\perp] \, \widehat D \, [\widehat V, \widehat V_\perp]^\top$, where $\widehat U \in \R^{d_1 \times r}, \widehat U_\perp \in \R^{d_1 \times (d_1-r)}, \widehat V \in \R^{d_2 \times r} \text{ and } \widehat V_\perp \in \R^{d_2 \times (d_2-r)}$. And we assume the SVD of the matrix $\Theta^*$ can be represented as $\Theta^* = UDV^\top$ where $U \in \R^{d_1 \times r}$ and $V \in \R^{d_2 \times r}$.
To transform the original generalized matrix bandits into generalized linear bandit problems, we follow the works in \cite{jun2019bilinear} and penalize those covariates that are complementary to $\widehat U$ and $\widehat V$. Specifically, we could orthogonally rotate the parameter space $\mathit{\Theta}$ and the action set $\mathcal{X}$ as:
\begin{align}
    \mathit{\Theta}^{\prime} = [\widehat U, \widehat U_\perp]^\top \mathit{\Theta} [\widehat V, \widehat V_\perp], \quad \mathcal{X}^\prime = [\widehat U, \widehat U_\perp]^\top \mathcal{X} [\widehat V, \widehat V_\perp],  \nonumber
\end{align} 
\setlength{\textfloatsep}{9pt}

Define the total dimension $p \coloneqq d_1d_2$, the effective dimension $k \coloneqq d_1d_2 - (d_1-r)(d_2-r)$ and the $r$-th largest singular value for $\Theta^*$ as $D_{rr}$, and vectorize the new arm space $\mathcal{X}^\prime$ and admissible parameter space as shown in Eqn. \eqref{rotate_arm} and \eqref{rotate_theta}. Then for the true parameter $\theta^*$ after transformation, we know that  $\theta^*_{k+1:p} = \vecc{\Theta^{*, \, \prime}_{r+1:d_1,r+1:d_2}}$ is almost null based on results in~\cite{stewart1990matrix} and Theorem \ref{thm_rscbound}:
\begin{align}
    \norm{\theta^*_{k+1:p}}_2 &= \norm{\widehat U_\perp^\top U D V^\top \widehat V_\perp}_F \leq \norm{\widehat U_\perp^\top U}_F \norm{\widehat V_\perp^\top V}_F \cdot\norm{D}_{\text{op}} \lesssim \dfrac{(d_1+d_2)Mr}{T_1 D_{rr}^2}\log\left(\frac{d_1+d_2}{\delta}\right) \coloneqq S_\perp. \label{Sperp}
\end{align}
Therefore, this problem degenerates to an equivalent $d_1d_2-$dimensional generalized linear bandit with a sparse structure (i.e. last $p-k$ entries of $\theta^*$ are almost null according to Eqn. \eqref{Sperp}). To reload the notation we define $\mathcal{X}_0, \mathit{\Theta}_0$ as the new feature set and parameter space as shown in Algorithm \ref{alg_estt}. 
\begin{algorithm}[t]
\caption{LowGLM-UCB} \label{alg1}
\begin{algorithmic}
\State {\bfseries Input:} $T_2,k,\mathcal{X}_0$, the probability rate $\delta$, penalization parameters $(\lambda_0,\lambda_\perp)$.
\State Initialize $M_1(c_\mu) = \sum_{i=1}^{T_1} x_{s_1,i} \, x_{s_1,i}^\top + \Lambda/c_\mu$. 
\For{ $t \geq 1$}
\State Estimate $\hat \theta_t$ according to \eqref{hattheta}. 
\State Choose the arm $x_t = \arg\max_{x \in \mathcal{X}_0} \{\mu({x}^\top{\hat \theta_t}) + \rho_{t}(\delta)\norm{x}_{M_t^{-1}(c_\mu)} \}$, receive $y_t$,
\State Update $M_{t+1}(c_\mu) \longleftarrow M_t(c_\mu) + x_t x_t^\top$.
\EndFor
\end{algorithmic}
\end{algorithm}
\hspace{-2.3 mm} 

\textbf{Remark.} Note the magnitude of $D_{rr}$ would be free of $d$ since $\Theta^*$ contains only $r$ nonzero singular values, and hence we assume that $D_{rr} = \Theta(1/\sqrt{r})$ under Assumption \ref{assu_bound}. This issue has been ignored in all previous analysis of explore-then-commit-type algorithms (e.g. ESTR~\cite{jun2019bilinear}, LowESTR~\cite{lu2021low}), where the final regret bound of them should be of order $\tilde{O} (d^{3/2}r\sqrt{T})$ instead of the originally-used $\tilde{O} (d^{3/2}\sqrt{rT})$ because of the existence of $D_{rr}$. 

\subsection{Stage 2 of G-ESTT}\label{subsec:gestt}
After reducing the original generalized matrix bandit problem into an identical $p$-dimensional generalized linear bandit problem in stage 2, we can reformulate the problem in the following way: at each round $t$, the agent chooses a vector $x_t$ of dimension $p$ from the transformed action set $\mathcal{X}_0$, and observes a noisy reward $y_t = \mu({x_t}^\top{\theta^*}) + \eta_t$. To make use of our additional knowledge shown in Eqn. \eqref{Sperp}, we propose LowGLM-UCB as an extension of the standard generalized linear bandit algorithm GLM-UCB~\citep{filippi2010parametric} combined with self-normalized martingale technique~ \citep{abbasi2011improved}. Specifically, we consider the following maximum quasi-likelihood estimation problem shown in Eqn. \eqref{est_theta} for each round with a weighted regularizer, where the regularizer is $ \norm{\theta}_{\Lambda}^2/2= \theta^{\top} \Lambda \theta/2$ for some positive definite diagonal matrix $\Lambda = \text{diag}(\lambda_0, \dots, \lambda_0, \plambda, \dots, \plambda)$ with $\lambda_0$ only applied to the first $k$ diagonal entries. By enlarging $\lambda_\perp$, we ensure more penalization forced on the last $p-k$ element of $\theta^*$ as desired.
\begin{gather}
    \hat \theta_t = \arg\max_\theta \widetilde{L}_t^{\Lambda}(\theta), \nonumber \\
    \widetilde{L}_t^{\Lambda}(\theta) = \sum_{i=1}^{T_1} \left[ y_{s_1,i} {x_{s_1,i}}^\top{\theta} - b({x_{s_1,i}}^\top{\theta}) \right]+
    \sum_{i=1}^{t-1} \left[ y_i {x_i}^\top{\theta} - b({x_i}^\top{\theta}) \right]-\frac{1}{2} \norm{\theta}_{\Lambda}^2. \label{est_theta}
\end{gather}\vspace{-3 mm}

Here $x_{s_1,i}$ in Eqn.~\eqref{est_theta} is the special vectorization shown in Eqn. \eqref{rotate_arm} of ${[\widehat U, \widehat U_\perp]^\top X_i [\widehat V, \widehat V_\perp]}$ where $X_i$ is the arm we randomly pull at $i$-th step in stage 1, and $y_{s_1,i}$ is the corresponding payoff we observe. $x_i$ in the second summation of Eqn.~\eqref{est_theta} refers to the arm we pull at $i$-th step in stage 2.
Since $\widetilde{L}_t^{\Lambda}(\theta)$ is a strictly concave function of $\theta$, we have its gradient equal to 0 at the maximum $\hat\theta_t$, i.e. $\nabla_\theta  \widetilde{L}_t^{\Lambda}(\theta) \vert_{\hat \theta_t} = 0$. In what follows, for $t \geq 2, \theta \in \mathbb{R}^p$ we define the function $g_t(\theta)$ and have that
\begin{gather}
   \nabla_\theta \widetilde{L}_t^{\Lambda}(\theta) = \sum_{i=1}^{T_1} y_{s_1,i} \, x_{s_1,i} + \sum_{i=1}^{t-1} y_i \, x_i - \bigg( \, \underbrace{\sum_{i=1}^{T_1} \mu({x_{s_1,i}}^\top{\theta}) x_{s_1,i} + \sum_{i=1}^{t-1} \mu({x_i}^\top{\theta}) x_i + \Lambda \theta}_{
    \textstyle
    \begin{gathered}
       \coloneqq g_t(\theta)
    \end{gathered}} \, \bigg), \nonumber \\ 
    \nabla_\theta  \widetilde{L}_t^{\Lambda}(\theta) \vert_{\hat \theta_t} = 0  \quad \Longrightarrow \quad g_t(\hat \theta_t) = \sum_{i=1}^{T_1} y_{s_1,i} \, x_{s_1,i} + \sum_{i=1}^{t-1} y_i \, x_i. \label{hattheta}
\end{gather}
We also define a matrix function $M_t(s) = \sum_{i=1}^{T_1} x_{s_1,i} x_{s_1,i}^\top + \sum_{k=1}^{t-1} x_k x_k^\top + \Lambda / s$ for $s \in \R^+$ and denote $V_t \coloneqq M_t(1)$. Furthermore, a remarkable benefit of reusing the actions $\{X_i\}_{i=1}^{T_1}$ we randomly pull in stage 1 is that they contain more randomness and are preferable to the ones we select based on some strategy in stage 2 regarding the parameter estimation because most vector recovery theory requires sufficient randomness during sampling. More inspiring, the projection step in the tradition GLM-UCB~\cite{filippi2010parametric}, which might be nonconvex and hence hard to solve, is no longer required due the consistency of $\hat \theta_t$ after reutilizing $\{X_i\}_{i=1}^{T_1}$. Specifically, if we assume the true parameter  $\theta^*$ lies in the interior of $\mathit{\Theta}_0$ and the sampling distribution $\mathcal{D}$ satisfies sub-Gaussian property with parameter $\sigma$, and Assumption \ref{assu_bound}, \ref{assu_link} held, then we can show that $\norm{\hat \theta_t - \theta^*}_2 \leq 1$ holds with probability at least $1-\delta$ as long as $T_1 \geq ((\hat{C}_1\sqrt{p}+\hat{C}_2\sqrt{\log(1/\delta))}/{\sigma^2})^2 + {2 B}/{\sigma^2}$ holds for some absolute constants $\hat{C}_1,\hat{C}_2$ with the definition $B \coloneqq 16\sigma_0^2(p+\log(1/\delta))/c_\mu^2$. An intuitive explanation along with a rigorous proof are deferred to Appendix \ref{sec:app:ucbglm} due to the space limit. The proposed LowGLM-UCB is shown in Algorithm \ref{alg1}, and its regret analysis is presented in Theorem \ref{thm_regalg1} in Appendix.

Notice that we can simply replace $M_t(c_\mu)$ by $V_t$ in Algorithm \ref{alg1}, and the regret bound would increase at most up to a constant factor  (Appendix \ref{sec:app:remark}). A potential drawback of Algorithm \ref{alg1} is that in each iteration we have to calculate $\hat\theta_t$, which might be computationally expensive. 
We could resolve this problem by only recomputing $\hat\theta_t$ whenever $\vert M_t(c_\mu)\vert$ increases significantly, i.e. by a constant factor $C > 1$ in scale. And consequently we only need to solve the Eqn. \eqref{hattheta} for $O(\log(T_2))$ times up to the horizon $T_2$, which remarkably saves the computation.  Meanwhile, the bound of the regret would only increase by a constant multiplier $\sqrt{C}$. We call this modified algorithm as PLowGLM-UCB with the initial letter ``P'' standing for ``Parsimonious''. Its pseudo-code and regret analysis are given in Appendix \ref{app:plow}. Equipped with LowGLM-UCB in stage 2, we deduce the overall regret of G-ESTT:

\subsubsection{Overall regret of G-ESTT} \label{subsec:overreg}
To quantify the performance of our algorithm, we first define $\alpha_t^x (\cdot)$ and $\beta_t^x (\cdot)$ as
\begin{align}
\alpha_t(\delta) &\coloneqq \dfrac{k_\mu}{c_\mu} \bigg(\sigma_0 \sqrt{k \log(1+ \frac{c_\mu S_0^2 t}{k\lambda_0})+ \frac{c_\mu S_0^2 t}{\lambda_\perp}-\log({\delta^2})}+ \sqrt{c_\mu}(\sqrt{\lambda_0}S_0+\sqrt{\lambda_\perp}S_\perp)\bigg),       \label{beta_def}  \\
    \beta_t^x (\delta)& \coloneqq \alpha_t(\delta) 
    \norm{x}_{M_t^{-1}(c_\mu)}. \label{alpha_def}
\end{align}
And the following Theorem \ref{thm_regestt} exhibits the overall regret bound for G-ESTT. 
\begin{algorithm}[t]
\caption{Generalized Explore Subspace Then Subtract (G-ESTS)} \label{alg_ests}
\begin{algorithmic}[1]
\Input $\mathcal{X},T,T_1,\mathcal{D}$, the probability rate $\delta$, parameters for Stage 2: $\lambda, \lambda_\perp$. \vspace{0.07 cm}
\Stage \textbf{1: Subspace Estimation}
\State Randomly choose $X_t \in \mathcal{X}$ according to $\mathcal{D}$ and record $X_t, Y_t$ for $t = 1,\dots T_1$.
\State Obtain $\widehat \Theta$ from Eqn. \eqref{loss}, and calculate its full SVD as $\widehat \Theta = [\widehat U, \widehat U_\perp] \, \widehat D \, [\widehat V, \widehat V_\perp]^\top$ where $\widehat U \in \mathbb{R}^{d_1 \times r}, \widehat V \in \mathbb{R}^{d_2 \times r}$. \vspace{0.07 cm}
\Stage \textbf{2: Low Dimensional Bandits} 
\State Rotate the arm feature set: $\mathcal{X}^\prime \coloneqq [\widehat U, \widehat U_\perp]^\top \mathcal{X} [\widehat V, \widehat V_\perp]$ and the admissible parameter space: $\mathit{\Theta}^{\prime} \coloneqq [\widehat U, \widehat U_\perp]^\top \mathit{\Theta} [\widehat V, \widehat V_\perp]$.
\State Define the vectorized arm set so that the last $(d_1-r) \cdot (d_2-r)$ components are negligible, and then \textbf{drop} them:
\begin{align}
    \mathcal{X}_{0, sub} \coloneqq \{\vecc{\mathcal{X}^\prime_{1:r,1:r}},\vecc{\mathcal{X}^\prime_{r+1:d_1,1:r}},  \vecc{\mathcal{X}^\prime_{1:r,r+1:d_2}}\}, \label{drop_arm}
\end{align}
and also refine the parameter set accordingly:
\begin{align}
    \mathit{\Theta}_{0,sub} \coloneqq \{\vecc{\mathit{\Theta}^\prime_{1:r,1:r}},\vecc{\mathit{\Theta}^\prime_{r+1:d_1,1:r}}, \vecc{\mathit{\Theta}^\prime_{1:r,r+1:d_2}}\}. \label{drop_theta}
\end{align}
\State For $T_2 = T -T_1$ rounds, invoke any misspecified generalized linear bandit algorithm with $\mathcal{X}_{0,sub},\mathit{\Theta}_{0,sub},k = (d_1+d_2)r - r^2$.
\end{algorithmic}
\end{algorithm}
\begin{theorem} \textup{(Regret of G-ESTT)}\label{thm_regestt}
Suppose we set $T_1 \asymp \sqrt{M(d_1+d_2)rT\log((d_1+d_2)/\delta)}/D_{rr}$, and we invoke LowGLM-UCB (or PLowGLM-UCB) in stage 2 with $\rho_t(\delta) = \alpha_{t+T_1}(\delta/2), p = d_1d_2, k = (d_1+d_2)r-r^2$, $\lambda_\perp = {c_\mu S_0^2 T}/(k \log(1+{c_\mu S_0^2 T}/({k \lambda_0})))$, and the rotated arm sets $\mathcal{X}_0$ and available parameter space $\mathit{\Theta}_0$. With $M = O((d_1+d_2)^2)$, the overall regret of G-ESTT is, with probability at least $1 - \delta$,
\begin{align}
    &Regret_T =  \tilde{O} \left((\frac{\sqrt{r(d_1+d_2)M}}{D_{rr}} + k) \sqrt{T} \right)
\nonumber 
\end{align} 
Specifically, with $M = O((d_1+d_2)^2)$, the regret bound becomes $\tilde{O} \left(({\sqrt{r(d_1+d_2)^3}}/{D_{rr}} + k) \sqrt{T} \right)$.
\end{theorem}

\subsection{Stage 2 of G-ESTS}\label{subsec:gests}
Although G-ESTT is more efficient than all existing algorithms on our problem setting, it still needs to calculate the MLE in high dimensional space which might be increasingly formidable with large sizes of feature matrices. Note this computational issue remains ubiquitous among most bandit algorithms on high dimensional problems with sparsity, not to mention these algorithms rely on multiple unspecified hyperparameters. Therefore, to handle this practical issue, we propose another fast and efficient framework called G-ESTS in this section.

Inspired by the success of dimension reduction in machine learning~\cite{van2009dimensionality}, we propose G-ESTS as shown in Algorithm \ref{alg_ests}. And we summarize the core idea of G-ESTS as: After rearranging the vectorization of the action set $\mathcal{X}^\prime$ and the unknown $\Theta^{\prime\, *}$ as we have shown in Eqn. \eqref{rotate_arm} and \eqref{rotate_theta} for G-ESTT, we can simply exclude, rather than penalize, the subspaces that are complementary to the rows and columns of $\widehat\Theta$. In other words, we could remove the last $p-k$ entries directly, i.e. Eqn. \eqref{drop_arm} and \eqref{drop_theta}. Intriguingly, not only can we get a low-dimensional ($k$) generalized linear bandit problem in stage 2, where redundant dimensions are excluded and hence any state-of-the-art algorithms could be readily invoked. Specifically, by utilizing any misspecified generalized linear bandit algorithm, we could validate the following Theorem~\ref{thm_regests}.



\subsubsection{Overall regret of G-ESTS}
\begin{theorem} \textup{(Regret of G-ESTS)}\label{thm_regests}
Suppose we set $T_1 \asymp \sqrt{M(d_1+d_2)^{3/2}r^{3/2}T\log((d_1+d_2)/\delta)}/D_{rr}$, and we invoke any efficient misspecified generalized linear bandit algorithm with regret bound $\tilde{O}(\epsilon \sqrt{k} T)$ \footnote{Modern misspecified generalized linear bandit algorithms can achieve $\tilde{O}(\epsilon \sqrt{k} T)$ bound of regret where $\epsilon$ is the misspecified rate.} in stage 2 with $p = d_1d_2, k = (d_1+d_2)r-r^2$, and the reduced arm sets $\mathcal{X}_{0,sub}$ and available parameter space $\mathit{\Theta}_{0,sub}$. The overall regret of G-ESTS is, with probability at least $1 - \delta$, 
\begin{align}
    &Regret_T = \widetilde{O} \left((\frac{\sqrt{r^{3/2}(d_1+d_2)^{3/2}M}}{D_{rr}} + k) \sqrt{T} \right).
    \nonumber
\end{align} 
Specifically, with $M = O((d_1+d_2)^2)$, the regret bound becomes $\widetilde{O}\left({\sqrt{r^{3/2}(d_1+d_2)^{7/2}T}}/{D_{rr}}\right)$.
\end{theorem}

Although our G-ESTS could achieve decent theoretical regret bound only equipped with misspecified generalized linear bandit algorithms, we showcase in practice it can work well with any state-of-the-art generalized linear bandit algorithm: In the following experiments, We will implement the SGD-TS algorithm~\citep{ding2021efficient} in stage 2 of G-ESTS since SGD-TS could efficiently proceed with only $O(dT)$ complexity for $d-$dimensional features over $T$ rounds. Therefore, the total computational complexity of stage 2 is at most $O(T_2(d_1+d_2)r)$, which is significantly less than that of other methods for low-rank matrix bandits (e.g. LowESTR~\citep{lu2021low}). And the total time complexity of G-ESTS would only scale ${O}(T_1d_1d_2/\epsilon^2+T_2(d_1+d_2)r)$ where $\epsilon$ is the accuracy for subgradient methods in stage 1. This fact also firmly validates the practical superiority of our G-ESTS approach. We naturally believe that this G-ESTS framework can be easily 
implemented in the linear setting as a special case of GLM, where in stage 2 one can utilize any linear bandit algorithm accordingly. In addition, we can easily modify our approaches for the contextual setting by merely transforming the action sets at each iteration with the same regret bound. More details with pseudo-codes for the contextual case are in Appendix \ref{context}.
\section{Experiments}\label{sec::exp}
In this section, we show by simulation experiments that our proposed G-ESTT (with LowGLM-UCB), G-ESTS (with SGD-TS) outperform existing algorithms for the generalized low-rank matrix bandit problems. Since we are the first to propose a practical algorithm for this problem, currently there is no existing literature for comparison. In order to validate the advantage of utilizing low-rank structure and generalized reward functions, we compare with the original SGD-TS after na\"ively flattening the $d_1$ by $d_2$ matrices without using the low-rank structure, and LowESTR~\citep{lu2021low}, which works well for linear low-rank matrix bandits.

\par
\begin{figure}[t]
\begin{minipage}[b]{0.245\linewidth}
    \centering
    \includegraphics[width = 0.95\textwidth]{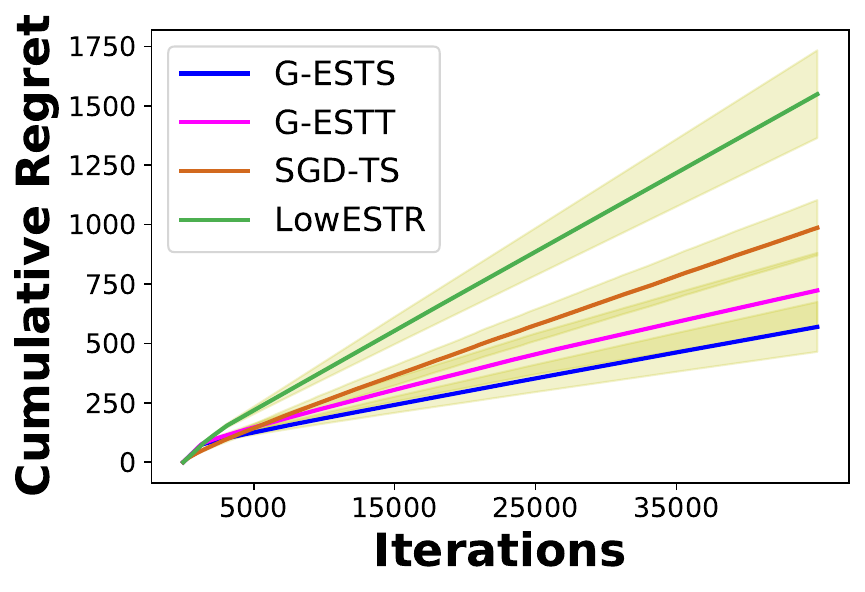}
    \vspace{-1mm}
    \captionof*{figure}{(a)}
\end{minipage}
\begin{minipage}[b]{0.245\linewidth}
    \centering
    \includegraphics[width = 0.95\textwidth]{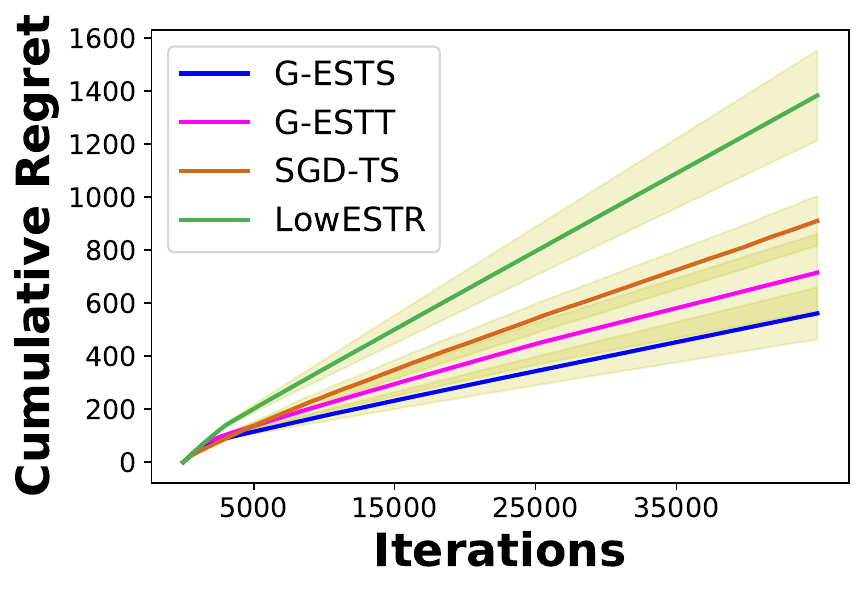}
    \vspace{-1mm}
    \captionof*{figure}{(b)}
\end{minipage}
\begin{minipage}[b]{0.245\linewidth}
    \centering
    \includegraphics[width = 0.95\textwidth]{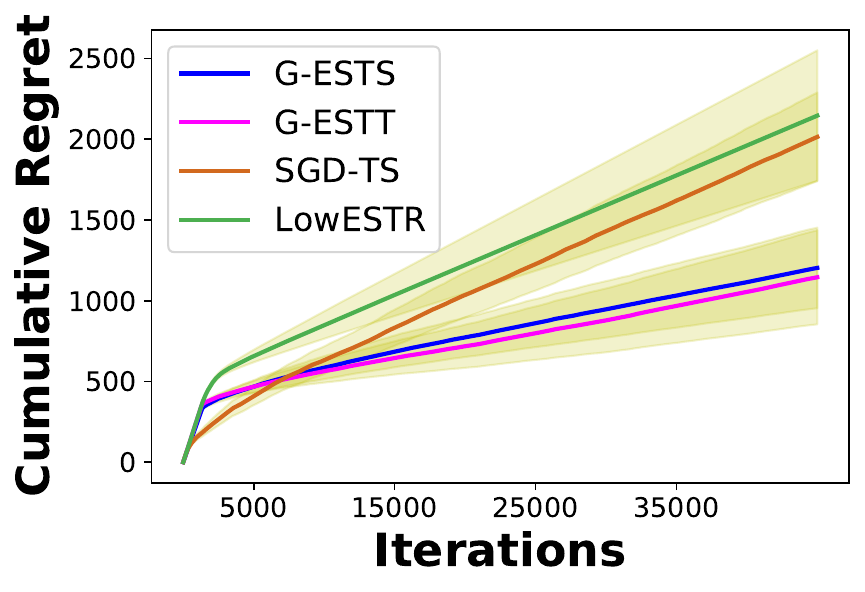}
    \vspace{-1mm}
    \captionof*{figure}{(c)}
\end{minipage}
\begin{minipage}[b]{0.245\linewidth}
    \centering
    \includegraphics[width = 0.95\textwidth]{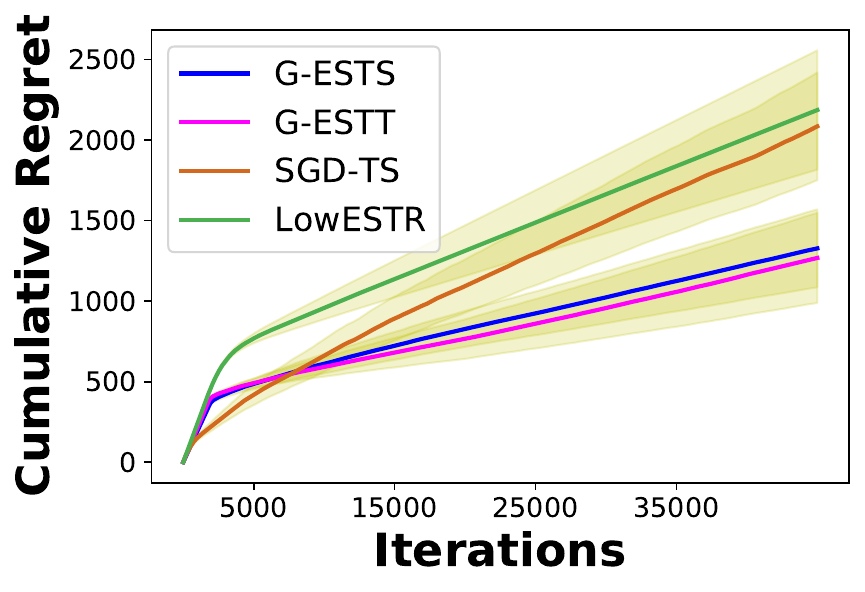}
    \vspace{-1mm}
    \captionof*{figure}{(d)}
\end{minipage}
\vspace{-3.5mm}
\caption{Plots of regret curves of algorithm G-ESTS, G-ESTT, SGD-TS and LowESTR under four settings ($480$ arms). (a): diagonal $\Theta^*$ $d_1=d_2=10,r=1$; (b): diagonal $\Theta^*$ $d_1=d_2=12,r=1$; (c): non-diagonal $\Theta^*$ $d_1=d_2=10,r=2$; (d): non-diagonal $\Theta^*$ $d_1=d_2=12,r=2$.}
\label{exp}
\vspace{-0.1 cm}
\end{figure}

 We simulate a dataset with $d_1=d_2 = 10 \,(12)$ and $r = 1 \, (2)$: when $r = 1$, we set the diagonal matrix $\Theta^*$ as diag$(\Theta^*) = (0.8,0,\cdots,0)$. When $r = 2$, we set $\Theta^*=v_1v_1^\top+v_2v_2^\top$ for two random orthogonal vectors $v_1, v_2$ with $\norm{v_1}_2 = \norm{v_2}_2 = 3$. For arms we draw $480 \, (1000)$ random matrices from $\{X \in \mathbb{R}^{d_1 \times d_2}: \, \norm{X}_F \leq 1\}$, and we build a logistic model where the payoff $y_t$ is drawn from a Bernoulli distribution with mean $\mu(X_t^\top \theta^*)$. More details on the hyper-parameter tuning are in Appendix \ref{sec:app:exp}. Each experiment is repeated 100 times for credibility and the average regret, along with standard deviation, is displayed in Figure \ref{exp}. Note that our experiments are more comprehensive 
than those in~\citep{lu2021low}. And due to the expensive time complexity of UCB-based baselines (Table \ref{table:time1}), it is 
formidable for us to increase $d$ here. 
\begin{table}[t]
\caption{Time in minutes required to make decisions all over round $T$ in simulations ($480$ arms).}
\label{table:time1}
\vskip 0.06in
\begin{center}
\begin{small}
\begin{sc}
\begin{tabular}{l|cc|cc}
\toprule
$d$ & \multicolumn{2}{c|}{10}&\multicolumn{2}{c}{12} \\
\midrule
$r$ &1&2&1&2 \\
\midrule
G-ESTS&\cellcolor{gray!30} 39.46 &\cellcolor{gray!30} 45.41&\cellcolor{gray!30} 41.28 &\cellcolor{gray!30} 48.52 \\
G-ESTT& 516.14& 531.95 &520.25 & 539.83 \\
SGD-TS& 99.57&101.34 &101.82 &104.42 \\
LowESTR&401.88& 419.15 &410.31 & 425.92 \\
\bottomrule
\end{tabular}
\end{sc}
\end{small}
\end{center}
\vskip -0.02in
\end{table}

\par From the plots, we observe that our algorithms G-ESTT and G-ESTS always achieve less regret compared with LowESTR and SGD-TS in all four scenarios consistently. Intriguingly, in the warm-up period SGD-TS incurs less regret compared with our methods due to the sacrifice of random sampling in stage 1, but our proposed framework quickly overtakes SGD-TS after utilizing the low-rank structure as desired. This phenomenon exactly coincides with our theory. Notice that G-ESTT is slightly better than G-ESTS in the case for $r=2$ especially in the very beginning of stage 2, and we believe it is because that our G-ESTT could reutilize the actions in stage 1 and hence could yield more robust performance when switching to stage 2. However, G-ESTS would gradually catch up with G-ESTT in the long run as expected. Besides, it costs G-ESTS extremely less running time than other existing methods to update the decisions due to its dimensional reduction as shown in Table \ref{table:time1}. We also observe that the cumulative regret of G-ESTS tends to become better eventually if we increase $T_1$ decently. (Further investigation and plots for $1000$ arms are in Appendix \ref{sec:app:exp}.) Moreover, to pre-check the efficiency of our Stein's lemma-based method for subspace estimation shown in Eqn. \eqref{loss}, we also tried some other low-rank subspace detection algorithms for comparison. The details are also deferred to Appendix \ref{app:compare_lowrank} due to the space limit.

\section{Conclusion}
In this paper, we discussed the generalized linear low-rank matrix bandit problem. We proposed two novel and efficient frameworks called G-ESTT and G-ESTS, and these two methods could achieve $\tilde{O}(\sqrt{M(d_1+d_2) rT}/D_{rr})$ and $\tilde{O}({\sqrt{MT}(d_1+d_2)^{3/4} r^{3/4}}/D_{rr})$ bound of regret under some mild conditions. The practical superiority of our proposed frameworks is also validated under comprehensive experiments.
\par There are several directions for our future work. Firstly, although the GLM has achieved tremendous success in various settings, recently some other models~\citep{zhou2020neural} that are proved to be powerful were proposed, so we can study these modern frameworks in the low-rank matrix case. 
Secondly, it seems more reasonable to continuously update the subspace estimation via online SVD over the total time horizon~\citep{jun2019bilinear} on-the-fly with a randomized strategy. Therefore, we can consider something like arbitrarily choosing an admissible subset of the action set at each round~\citep{schumann2019group} without hurting the regret too much.


\bibliographystyle{plain}
\bibliography{arxiv_mod}

\clearpage
\newpage
\appendix
\section*{Appendix}

\section{Clarification about $\sigma_0^2$}
It is a common assumption that the random noise $\eta_t$ in Eqn. \eqref{subgform} is a sub-Gaussian random variable in GLM, and here we would briefly explain this assumption.
\begin{lemma} \label{lemma_glmsubg} \textup{(Sub-Gaussian property for GLM residuals)}
For any generalized linear model with a probability density function or probability mass function of the canonical form
\begin{align}
    f(Y=y;\theta,\phi) = \exp\left(\frac{y \theta - b(\theta)}{\phi} + c(y,\phi) \right), \nonumber
\end{align}
where the function $b(\cdot)$ is Lipschitz with parameter $k_\mu$. Then we can conclude that the random variable $(Y - b^{\prime} (\theta))=(Y - \mu(\theta))$ satisfies sub-Gaussian property with parameter at most $\sqrt{\phi \, k_\mu}$.
\end{lemma}
\textit{Proof.} We prove the Lemma \ref{lemma_glmsubg} based on its definition directly. For any $t \in \R$, we have:
\begin{align}
    \E\left[\exp\{t\left(Y-b^{\prime}(\theta)\right)\}\right] &=  \int_{- \infty}^{+ \infty} \exp \left\{t(y-b^{\prime}(\theta))+\frac{y \theta - b(\theta)}{\phi} + c(y,\phi)\right\} \, dy  \nonumber \\
    &= \int_{- \infty}^{+ \infty} \exp \left\{\frac{(\theta + \phi t)y - b(\theta + \phi t)}{\phi} + c(y,\phi)\right\} \nonumber \\
    & \quad \quad \quad \quad  \times\exp\left\{\frac{b(\theta+\phi t) - b(\theta) - \phi t b^{\prime} (\theta)}{\phi}\right\} \, dy \nonumber \\
    &= \exp\left\{\frac{b(\theta+\phi t) - b(\theta) - \phi t b^{\prime} (\theta)}{\phi}\right\} \nonumber  \\
    &\stackrel{(i)}{=} \exp\left\{\frac{t^2 \phi \, b^{\prime \prime}(\theta + \delta \phi t)}{2} \right\} \leq \exp\left\{\frac{t^2 \, \phi \, k_\mu}{2}\right\}
    \coloneqq \exp\left\{\frac{t^2 \, \sigma_{0}^2}{2}\right\}, \nonumber
\end{align}
where the equality (i) is based on the remainder of Taylor expansion.  \hfill \qedsymbol
\newline This theorem tells us that it is a standard assumption that the noise $\eta_t$ in Eqn. \eqref{subgform} is a sub-Gaussian random variable. For instance, if we assume the inverse link function $\mu(\cdot)$ is globally Lipschitz with parameter $k_\mu$, we can simply take $\sigma_0^2 = k_\mu\phi$. And this assumption also widely holds under a class of GLMs such as the most popular Logistic model.

\section{Proof of Theorem \ref{thm_rscbound}} \label{app:stein}
\subsection{Useful Lemmas}
\begin{lemma}\textup{(Sub-guassian moment bound)} \label{lem:subgvar}
For sub-Gaussian random variable $X$ with parameter $\sigma^2$, i.e.
$$\mathbb{E}(\exp(sX)) \leq \exp\left(\frac{\sigma^2s^2}{2}\right), \quad \; \forall s \in \mathbb{R}.$$
Then we have $\text{Var}(X) = \mathbb{E}(X^2) \leq 4 \sigma^2$.
\end{lemma}
\textit{Proof.} It holds that,
\begin{align*}
    \mathbb{E}(X^2) &= \int_0^{+\infty} P(X^2 > t) \, dt \\
    &=\int_0^{+\infty} P(|X| > \sqrt{t}) \, dt \\
    &\leq 2 \int_0^{+\infty} \exp{(\frac{-t^2}{4\sigma^2})} \, dt \\
    &= 4 \sigma^2 \int_0^{+\infty} e^{-u}  \, du, \quad \ u=t/(2\sigma^2) \\
    &= 4 \sigma^2
\end{align*}
\hfill \qedsymbol
\begin{lemma}(\textup{Generalized Stein's Lemma,} \citep{diaconis2004use})\label{lem:stein} For a random variable $X$ with continously differentiable density function $p : \mathbb{R}^d \rightarrow \R$, and any continuously differentiable function $f:\mathbb{R}^d \rightarrow \R$. If the expected values of both $\nabla f(X)$ and $f(X)\cdot S(X)$ regarding the density $p$ exist, then they are identical, i.e.
$$\E[f(X) \cdot S(X)] = \E[\nabla f(X)].$$
\end{lemma}
This is a very famous result in the area of Stein's method, and we would omit its proof.
\begin{lemma}(\cite{minsker2018sub}) \label{lem:minsker} Let $Y_1, \dots, Y_n \in \mathbb{R}^{d_1 \times d_2}$ be a sequence of independent real random matrices, and assume that
$$\sigma_n^2 \geq \max\left(\opnorm{\sum_{j=1}^n \E(Y_jY_j^\top)},\opnorm{\sum_{j=1}^n \E(Y_j^\top Y_j)}\right).$$
Then for any $t \in \R^+$ and $\nu \in \R^+$, it holds that,
$$P\left(\opnorm{\sum_{j=1}^n \tilde{\psi}_\nu(Y_j) - \sum_{j=1}^n \E(Y_j)} \geq t \sqrt{n}\right) \leq 2(d_1+d_2) \exp\left(\nu t \sqrt{n} + \frac{\nu^2 \sigma_n^2}{2}\right)$$
\end{lemma}
The detailed proof of this lemma is based on a series of work proposed in~\citep{minsker2018sub}. And we would omit it here as well. Based on Lemma \ref{lem:stein} and \ref{lem:minsker}, we would propose the following Lemma \ref{lem:krishna} adapted from the work in~\citep{yang2017high}. And this Lemma serves as a crux for the proof of Theorem \ref{thm_rscbound}.

\begin{lemma}\label{lem:krishna} $L:\R^{d_1 \times d_2} \rightarrow \R$ is the loss function defined in Eqn. \eqref{loss}. Then by setting 
\begin{gather*}
    t = \sqrt{2(d_1+d_2)M(4\sigma_0^2+S_f^2)\log\left(\frac{2(d_1+d_2)}{\delta}\right)}, \\ \nu  =\frac{t}{(4\sigma_0^2+S_f)M(d_1+d_2)\sqrt{T_1}}= \sqrt{\dfrac{2\log\left(\frac{2(d_1+d_2)}{\delta}\right)}{T_1(d_1+d_2)M(4\sigma_0^2+S_f^2)}},
\end{gather*}
we have with probability at least $1-\delta$, it holds that
$$P\left(\opnorm{\nabla L(\mu^*\Theta^*)} \geq \frac{2t}{\sqrt{T_1}}\right) \leq \delta,$$
where $\mu^* = \E[\mu^\prime (\inp{X}{\Theta^*})] \geq c_\mu > 0$.
\end{lemma}
\textit{Proof.} Based on the definition of our loss function $L(\cdot)$ in Eqn. \eqref{loss}, we have that
\begin{align*}
    \nabla_x L(\mu^*\Theta^*) &= 2\mu^*\Theta^* - \frac{2}{T_1}\sum_{i=1}^{T_1} \tilde{\psi}_\nu (y \cdot S(x))  \\
    &= 2\E[\mu^\prime(\inp{X_1}{\Theta^*})]\Theta^* - \frac{2}{T_1}\sum_{i=1}^{T_1} \tilde{\psi}_\nu (y_i \cdot S(X_i)) \\
    &\stackrel{\textup{(i)}}{=} 2\E[\mu(\inp{X_1}{\Theta^*})S(X_1)] - \frac{2}{T_1}\sum_{i=1}^{T_1} \tilde{\psi}_\nu (y_i \cdot S(X_i)) \\
    &\stackrel{\textup{(ii)}}{=} 2\left[\E(Y_1\cdot S(X_1)) -\frac{1}{T_1}\sum_{i=1}^{T_1} \tilde{\psi}_\nu (y_i \cdot S(X_i)) \right]
\end{align*}
where we have (i) due to the generalized Stein's Lemma (Lemma \ref{lem:stein}), and (ii) comes from the fact that the random noise $\eta_1 = y_1 -\mu(\inp{X_1}{\Theta^*})$ is zero-mean and independent with $X_1$. Therefore, in order to implement the Lemma \ref{lem:minsker}, we can see that it suffices to get $\sigma^2$ defined as:
$$\sigma^2 = \max\left(\opnorm{\sum_{j=1}^n \E[y_j^2 S(X_j)S(X_j)^\top]},\opnorm{\sum_{j=1}^n \E[y_j^2 S(X_j)^\top S(X_j)]}\right).$$
It holds that,
\begin{align*}
    \opnorm{\sum_{j=1}^{T_1} \E[y_j^2 S(X_j)S(X_j)^\top]} &\leq T_1 \times \opnorm{\E[y_1^2 S(X_1)S(X_1)^\top]} \\
    &\hspace{-1 cm}= T_1\times \opnorm{\E[(\eta_1 + \mu(\inp{X_1}{\Theta^*}))^2 S(X_1)S(X_1)^\top]} \\
    &\hspace{-1 cm}= T_1\times \opnorm{\E[\eta_1^2 S(X_1)S(X_1)^\top] + \E[\mu(\inp{X_1}{\Theta^*}))^2 S(X_1)S(X_1)^\top]} \\
    &\hspace{-1 cm}=  T_1\times \opnorm{\E(\eta_1^2)\E[ S(X_1)S(X_1)^\top] + \E[\mu(\inp{X_1}{\Theta^*}))^2 S(X_1)S(X_1)^\top]} \\
    &\hspace{-1 cm}\stackrel{\textup{(i)}}{\leq} T_1 \times \opnorm{4\sigma_0^2 \, \E[ S(X_1)S(X_1)^\top] + S_f^2 \, \E[ S(X_1)S(X_1)^\top]} \\
    &\hspace{-1 cm}= (4\sigma_0^2+S_f^2)T_1 \times \opnorm{\E[ S(X_1)S(X_1)^\top]}
\end{align*}
where the inequality (i) comes from the fact that $|\mu(\inp{X_1}{\Theta^*})| \leq S_f$, and $S(X_1)S(X_1)^\top$ is always positive semidefinite. Next, without loss of generality we assume $X_i$ are independent across rows, since if $X_i$ are independent across columns we can study the value of $\opnorm{\E[S(X_1)^\top S(X_1)]}$ given the fact that the largest singular values of $S(X_1)^\top S(X_1)$ and $S(X_1)S(X_1)^\top$ are identical for arbitrary $X_1$. We know that $\E [S(X_1)S(X_1)^\top]$ is always symmetric and positive semidefinite, and hence we have for any $u \in \R^{d_1}$ with $\|u\| = 1$
\begin{align*}
	 u^\top {\E[ S(X_1)S(X_1)^\top]} u &= {\E[ u^\top S(X_1)S(X_1)^\top u]} = \E\left[ \norm{S(X_1)^\top u}^2 \right] \\
	 & = \sum_{j=1}^{d_2} \E\left[\left(\sum_{i=1}^{d_1} S_{i,j}(X_1) u_i \right)^2 \right] \\
	 & = \sum_{j=1}^{d_2}  \E\left[\sum_{i=1}^{d_1} S_{i,j}(X_1)^2 u_i^2 \right] \leq d_2 M,
\end{align*}
and this result implies that $\opnorm{\E[ S(X_1)S(X_1)^\top]} \leq d_2 M \leq (d_1+d_2)M$
Therefore, we have that 
$$\opnorm{\sum_{j=1}^{T_1} \E[y_j^2 S(X_j)S(X_j)^\top]} \leq (4\sigma_0^2 + S_f^2)(d_1+d_2)T_1M.$$
And similarly, we can prove that
$$\opnorm{\sum_{j=1}^{T_1} \E[y_j^2 S(X_j)^{\top}S(X_j)]} \leq (4\sigma_0^2 + S_f^2)(d_1+d_2)T_1M.$$
Therefore, we can take $\sigma^2 = (4\sigma_0^2 + S_f^2)(d_1+d_2)T_1M$ consequently. By using Lemma \ref{lem:minsker}, we have
$$P\left(\opnorm{\nabla L(\mu^* \Theta^*)} \geq \dfrac{2t}{\sqrt{T_1}}\right) \leq 2(d_1+d_2)\exp\left(-\nu t\sqrt{T_1} + \dfrac{\nu^2(4\sigma_0^2+S_f^2)M(d_1+d_2)T_1}{2}\right)$$
By plugging the values of $t$ and $\nu$ in Lemma \ref{lem:krishna}, we finish the proof. \hfill \qedsymbol
\subsection{Proof of Theorem \ref{thm_rscbound}}
Since the estimator $\widehat\Theta$ minimizes the regularized loss function defined in Eqn. \eqref{loss}, we have
$$L(\widehat\Theta) + \lambda_{T_1} \nunorm{\widehat\Theta} \leq L(\mu^*\Theta^*) + \lambda_{T_1} \nunorm{\mu^*\Theta^*}.$$
And due to the fact that $L(\cdot)$ is a quadratic function, we have the following expression based on multivariate Taylor's expansion:
$$L(\widehat\Theta) - L(\mu^*\Theta^*) = \inp{\nabla L(\mu^*\Theta^*)}{\Theta} + 2 \norm{\Theta}_F^2, \quad \ \text{where } \Theta = \widehat\Theta - \mu^*\Theta^*.$$
By rearranging the above two results, we can deduce that
\begin{align}
    2\norm{\Theta}_F^2 &\leq - \inp{\nabla L(\mu^*\Theta^*)}{\Theta} + \lambda_{T_1}\nunorm{\mu^*\Theta^*} - \lambda_{T_1} \nunorm{\widehat \Theta} \nonumber \\
    &\stackrel{\textup{(i)}}{\leq} \opnorm{\nabla L(\mu^*\Theta^*)} \nunorm{\Theta} + \lambda_{T_1}\nunorm{\mu^*\Theta^*} - \lambda_{T_1} \nunorm{\widehat \Theta}, \label{eq:krishnaineq}
\end{align}
where (i) comes from the duality between matrix operator norm and nuclear norm. Next, we represent the saturated SVD of $\Theta^*$ in the main paper as $\Theta^* = UDV^\top$ where $U \in \R^{d_1 \times r}$ and $V \in \R^{d_2 \times r}$, and here we would work on its full version, i.e.
$$\Theta^*= (U, U_\perp) \begin{pmatrix}
D & 0 \\
0 & 0
\end{pmatrix} (V, V_\perp)^\top = (U, U_\perp) D^* (V, V_\perp)^\top,$$
where we have $U_\perp \in \R^{d_1 \times (d_1 - r)}$, $D^* \in \R^{d_1 \times d_2}$ and $V_\perp \in \R^{d_2 \times (d_2 - r)}$. Furthermore, we define 
$$\Lambda = (U, U_\perp)^\top \Theta (V,V_\perp) = \begin{pmatrix}
U^\top \Theta V & U^\top \Theta V_\perp \\
U_\perp^\top \Theta V &
U_\perp^\top \Theta V_\perp 
\end{pmatrix} = \Lambda_1 + \Lambda_2$$
where we write
$$\Lambda_1  = \begin{pmatrix}
0 & 0 \\
0 & U_\perp^\top \Theta V_\perp 
\end{pmatrix}, \quad \Lambda_2 = \begin{pmatrix}
U^\top \Theta V & U^\top \Theta V_\perp \\
U_\perp^\top \Theta V &
0
\end{pmatrix}$$
Afterwards, it holds that
\begin{align}
\nunorm{\widehat\Theta} &= \nunorm{\mu^* \Theta^* + \Theta} = \nunorm{(U, U_\perp) (\mu^* D^* + \Lambda) (V, V_\perp)^\top}  \nonumber \\
&= \nunorm{\mu^*D^* + \Lambda} +\nunorm{\mu^*D^* + \Lambda_1 +\Lambda_2} \nonumber \\
&\geq \nunorm{\mu^*D^* + \Lambda_1} - \nunorm{\Lambda_2} \nonumber \\
&= \nunorm{\mu^*D}+ \nunorm{\Lambda_1} - \nunorm{\Lambda_2} \nonumber \\
&=\nunorm{\mu^*\Theta^*}+ \nunorm{\Lambda_1} - \nunorm{\Lambda_2}, \nonumber
\end{align}
which implies that 
\begin{align}
\nunorm{\mu^*\Theta^*} - \nunorm{\widehat\Theta} \leq \nunorm{\Lambda_2} - \nunorm{\Lambda_1} \label{eq:krishna1}
\end{align}
Combine Eqn. \eqref{eq:krishnaineq} and \eqref{eq:krishna1}, we have that
\begin{align}
    2 \norm{\Theta}_F^2 \leq \left(\opnorm{\nabla L(\mu^*\Theta^*)} + \lambda_{T_1} \right) \nunorm{\Lambda_2} + \left(\opnorm{\nabla L(\mu^*\Theta^*)} - \lambda_{T_1} \right) \nunorm{\Lambda_1} \nonumber
\end{align}
Then, we refer to the setting in our Lemma \ref{lem:krishna}, and we choose $\lambda = 4t/\sqrt{T_1}$ where the value of $t$ is determined in Lemma \ref{lem:krishna}, i.e.
$$\lambda_{T_1} = 4\sqrt{\frac{2(4\sigma_0^2+S_f^2)M(d_1+d_2)\log(2(d_1+d_2)/\delta)}{T_1}},$$
we know that $\lambda_{T-1} \geq 2 \opnorm{\nabla L(\mu^*\Theta^*)}$ with probability at least $1-\delta$ for any $\delta \in (0,1)$. Therefore, with probability at least $1-\delta$, we have
$$2\norm{\Theta}_F^2 \leq \frac{3}{2}\lambda_{T_1}\nunorm{\Lambda_2} - \frac{1}{2} \lambda_{T_1} \nunorm{\Lambda_1} \leq \frac{3}{2}\lambda_{T_1}\nunorm{\Lambda_2}$$.
Since we can easily verify that the rank of $\Lambda_2$ is at most $2r$, and by using Cauchy-Schwarz Inequality we have that $$2\norm{\Theta}_F^2 \leq \frac{3}{2} \lambda_{T_1} \sqrt{2r} \norm{\Lambda_2}_F \leq \frac{3}{2} \lambda_{T_1} \sqrt{2r} \norm{\Lambda}_F = \frac{3}{2} \lambda_{T_1} \sqrt{2r} \norm{\Theta}_F,$$
which implies that 
$$\norm{\Theta}_F \leq \frac{3}{4}\sqrt{2r} \lambda_{T_1} =  6\sqrt{\dfrac{(4\sigma_0^2+S_f^2)M(d_1+d_2)r\log(\frac{2(d_1+d_2)}{\delta})}{T_1}},$$
and it concludes our proof. \hfill \qedsymbol
\section{Theorem \ref{thm_regalg1} and its analysis}
\subsection{Theorem \ref{thm_regalg1}}
\begin{theorem} \label{thm_regalg1} \textup{(Regret of LowGLM-UCB)}
Under Assumption \ref{assu_bound} and \ref{assu_link}, for any fixed failure rate $\delta \in (0,1)$, if we run the LowGLM-UCB algorithm with $\rho_t(\delta) = \alpha_{t+T_1}(\delta/2)$ and
\begin{align}
    \lambda_\perp \asymp \dfrac{c_\mu S_0^2 T}{k \log(1+\frac{c_\mu S_0^2 T}{k \lambda_0})},  \nonumber
\end{align}
then the bound of regret for LowGLM-UCB $(Regret_{T_2})$ achieves ${\widetilde{O}(k \sqrt{T}+ T S_\perp)}$, with probability at least $1-\delta$.

\end{theorem}

\subsection{Proposition \ref{prop_mubound} with its proof}
We firstly present the following important Proposition \ref{prop_mubound} for obtaining the upper confidence bound.
\begin{proposition} \label{prop_mubound}
For any $\delta, t$ such that $\delta \in (0,1)$, $t \geq 2$, and for $\beta_t^x (\delta)$ defined in Eqn. \eqref{beta_def} and \eqref{alpha_def}, with probablity $1-\delta$, it holds that
\begin{align}
    \vert \mu(x^\top \theta^*) - \mu(x^\top \hat\theta_t)\vert \leq \beta_{t+T_1}^x(\delta), \label{prop_mubound_equ}
\end{align}
simultaneously for all $x \in \R$ and all $t \geq 2$.
\end{proposition}
\subsubsection{Technical Lemmas}
\begin{lemma} \label{lemma_goodprob} \textup{(Adapted from Abbasi-Yadkori et al., 2011, Theorem 1)}
Let $\{\mathcal{F}_t\}_{t=0}^\infty$ be a filtration and $\{x_t\}_{t=0}^\infty$ be an $\mathbb{R}^d$-valued stochastic process adapted to $\mathcal{F}_t$. Let $\{\eta_t\}_{t=0}^\infty$ be a real-valued stochastic process such that $\eta_t$ is adapted to $\mathcal{F}_t$ and is conditionally $\sigma_0$-sub-Gaussian for some $\sigma_0 > 0$, i.e.
\begin{align}
    \E[\exp(\lambda \eta_t) \vert \mathcal{F}_t] \leq \exp\left( \dfrac{\lambda^2 \sigma_0^2}{2} \right), \quad \quad \forall \lambda \in \R. \nonumber
\end{align}
Consider the martingale $S_t = \sum_{k=1}^{t} \eta_{k} x_k$ and the process $V_t = \sum_{k=1}^{t} x_{k} x_{k}^{\top} + \Lambda$ when $t \geq 2$. And $\Lambda$ is fixed and independent with sample random variables after time $m$. For any $\delta > 0$, with probability at least $1-\delta$, we have the following result simultaneously for all $t \geq m+1$:
\begin{align}
    \norm{S_t}_{V_t^{-1}} \leq \sigma_0 \sqrt{\log (\det(V_t)) - \log(\delta^2 \det(\Lambda))}. \nonumber
\end{align}
\end{lemma}
We defer the proof for this lemma to Section \ref{subsec:lemma_goodprob} since a lot of technical details are involved.

\begin{lemma} \label{lemma_inversepd} 
For any two symmetric positive definite matrix $A,B \in R^{p \times p}$ such that $A \preceq B$, we have $A B^{-1} A \preceq A$.
\end{lemma}
\proof Since $A \preceq B$ and both of them are invertible matrices, we have $B^{-1} \preceq A^{-1}$ directly based on positive definiteness property. Conjugate with $A$ on both sides we can directly obtain $A B^{-1} A \preceq A$. \hfill \qedsymbol

\begin{lemma} \label{lemma_valko} \textup{(Valko et al., 2014, Lemma 5)}
For any $T \geq 1$, let $V_{T+1} = \sum_{i=1}^{T} x_ix_i^{\top} + \Lambda \in R^{p}$ where $\Lambda = \text{diag}\{\lambda_1, \dots, \lambda_p\}$. And we assume that $\norm{x_i}_2 \leq S$. Then:
\begin{align}
    \log \dfrac{\vert V_{T+1} \vert}{\vert \Lambda \vert} \leq \max_{\{t_i\}_{i=1}^p} \sum_{i=1}^p \log \left(1 + \dfrac{S^2 t_i}{\lambda_i} \right), \nonumber
\end{align}
where the maximum is taken over all possible positive real numbers $\{t_i\}_{i=1}^p$ such that $\sum_{i=1}^p t_i = T$
\end{lemma}
\proof We aim to bound the determinant $\vert V_{T+1} \vert$ under the coordinate constrains $\norm{x_i}_2 \leq S$. Let's denote
\begin{align}
    U(x_1,\dots,x_T) = \lvert \Sigma + \sum_{t=1}^T x_t x_t^\top \rvert. \nonumber
\end{align}

Based on the property of the sum of rank-1 matrices (e.g. Valko et al., 2014, Lemma 4), we know that the maximum of $U(x_1,\dots,x_T)$ is reached when all $x_t$ are aligned with the axes:
\begin{align}
    U(x_1,\dots,x_T) = \max_{\substack{x_1, \dots, x_T; \\ x_t \in S \cdot \{e_1, \dots, e_N\}}} \lvert \Sigma + \sum_{t=1}^T x_t x_t^\top \rvert = \max_{\substack{t_1, \dots, t_N \text{positive integers}; \\ \sum_{i=1}^N t_i = T}} \lvert \text{diag}(\lambda_i + t_i)  \rvert \nonumber \\ 
    \leq \max_{\substack{t_1, \dots, t_N \text{positive integers}; \\ \sum_{i=1}^N t_i = S^2T}} \prod_{i=1}^N  (\lambda_i + S^2t_i). \nonumber 
\end{align} \hfill \qedsymbol

\subsubsection{Proof of Proposition \ref{prop_mubound}}
\proof Recall our definition of $g_t(\theta)$ and its gradient accordingly as 
\begin{align}
    g_t(\theta) &= \sum_{i=1}^{T_1} \mu({x_{s_1,i}}^\top{\theta}) x_{s_1,i} + \sum_{k=1}^{t-1} \mu(x_k^\top \theta) x_k + \Lambda \theta, \nonumber \\
    \nabla_\theta g_t(\theta) = \sum_{i=1}^{T_1} & \mu^\prime({x_{s_1,i}}^\top{\theta}) x_{s_1,i}x_{s_1,i}^\prime +\sum_{k=1}^{t-1} \mu^\prime (x_k^\top \theta) x_k x_k^\top + \Lambda \stackrel{\textup{(i)}}{\succeq}  c_\mu M_t(c_\mu), \label{gtgradientdef}
\end{align}
where the relation (i) holds if $\theta \in \mathit{\Theta}_0$. Based on Assumptions, we know the gradient $\nabla_\theta g_t(\theta)$ is continuous. Then the Fundamental Theorem of Calculus will imply that 
\begin{align}
    g_t(\theta^*) - g_t(\hat \theta_t) = G_t (\theta^* - \hat \theta_t), \nonumber
\end{align}
where
\begin{align}
    G_t = \int_0^1 \nabla_\theta g_t(s\theta^* + (1-s)\hat \theta_t) \, ds. \nonumber
\end{align}

Since we assume that the inverse link function $\mu(\cdot)$ is $k_\mu-$Lipshitz, and the matrix $G_t$ is always invertible due to the fact that at least we have $G_t \succeq \Lambda$, we can obtain the following result. Notice the inequality (i) comes from the fact that $G_t \succeq c_\mu M_t(c_\mu)$ and hence $M_t(c_\mu)^{-1}/c_\mu \succeq G_t^{-1}$.
\begin{align}
    \vert \mu(x^\top \theta^*) - \mu(x^\top \hat \theta_t) \vert &\leq k_\mu \vert x^\top (\theta^*-\hat \theta_t) \vert 
    = k_\mu \vert x^\top G_t^{-1} (g_t(\theta^*)-g_t(\hat \theta_t)) \vert \nonumber \\ 
    &\hspace{-12 mm}\leq k_\mu \norm{x}_{G_t^{-1}} \, \norm{g_t(\theta^*)-g_t(\hat \theta_t)}_{G_t^{-1}} \stackrel{\textup{(i)}}{\leq} 
    \frac{k_\mu}{c_\mu} \norm{x}_{M_t(c_\mu)^{-1}} \, \norm{g_t(\theta^*)-g_t(\hat \theta_t)}_{M_t(c_\mu)^{-1}}. \nonumber
\end{align}

In addition, based on the definition of $\hat \theta_t$ in Equation \eqref{hattheta}, we have $g_t(\hat \theta_t) - g_t(\theta^*) = \sum_{k=1}^{T_1} (y_{s_1,k} - \mu(x_{s_1,k}^\top \theta^*))x_{s_1,k} + \sum_{k=1}^{t-1} (y_k - \mu(x_k^\top \theta^*))x_k - \Lambda \theta^* =\sum_{k=1}^{T_1} \eta_{s_1,k}x_{s_1,k}+ \sum_{k=1}^{t-1} \eta_kx_k - \Lambda \theta^*$. Therefore, 
\begin{align}
    \vert \mu(x^\top \theta^*) - \mu(x^\top \hat \theta_t) \vert &\leq  \frac{k_\mu}{c_\mu} \norm{x}_{M_t(c_\mu)^{-1}} \, \norm{g_t(\hat \theta_t) - g_t(\theta^*)}_{M_t^{-1}(c_\mu)} \nonumber  \\
    &\hspace{-18 mm}\leq \frac{k_\mu}{c_\mu} \norm{x}_{M_t(c_\mu)^{-1}} \, \left(\norm{\sum_{k=1}^{T_1} \eta_{s_1,k}x_{s_1,k}+\sum_{k=1}^{t-1} \eta_k x_k}_{M_t^{-1}(c_\mu)} + \norm{\Lambda \theta^*}_{M_t^{-1}(c_\mu)}\right). \label{prop_pf_1}
\end{align}

Now, let's use Lemma \ref{lemma_goodprob} to bound the term $\norm{\sum_{k=1}^{T_1} \eta_{s_1,k}x_{s_1,k}+\sum_{k=1}^{t-1} \eta_k x_k}_{M_t^{-1}(c_\mu)}$. If we define the filtration $\mathcal{F}_t \coloneqq \{\{x_t,x_{t-1},\eta_{t-1},\dots,x_1,\eta_1\} \cup \{x_{s_1,k},\eta_{s_1,k}\}_{k=1}^{T_1}\}$, then for any $\delta \in (0,1)$, with probability $1-\delta$, it holds that for all $t \geq 2$,
\begin{align}
    \norm{\sum_{k=1}^{T_1} \eta_{s_1,k}x_{s_1,k}+\sum_{k=1}^{t-1} \eta_k x_k}_{M_t^{-1}(c_\mu)} \leq \sigma_0 \sqrt{\log \left( \dfrac{\vert M_t(c_\mu)\vert}{\vert  \frac{\Lambda}{c_\mu} \vert } \right) - 2 \log (\delta)}, \nonumber
\end{align}
where based on Lemma \ref{lemma_valko},
\begin{align}
  \log \left( \dfrac{\vert M_t(c_\mu)\vert}{\vert  \frac{\Lambda}{c_\mu} \vert } \right) & \leq \max_{\substack{t_i \geq 0, \\  \sum_{i=1}^t t_i = t+T_1}} \sum_{i=1}^p \log \left(1 + \dfrac{c_\mu S_0^2 t_i}{\lambda_i} \right) \nonumber \\ 
&\leq k \log \left(1+\frac{c_\mu S_0^2}{k \lambda_0}(t+T_1) \right) + (d-k) \log \left( 1 + \frac{c_\mu S_0^2}{(d-k) \lambda_\perp}(t+T_1)\right)\nonumber \\ &\leq  k \log \left(1+\frac{c_\mu S_0^2}{k \lambda_0}(t+T_1) \right) +  \frac{c_\mu S_0^2}{\lambda_\perp}(t+T_1). \label{prop_pf_2}
\end{align}

And next by Lemma \ref{lemma_inversepd}, we have
\begin{align}
    \norm{\Lambda \theta^*}_{M_t^{-1}(c_\mu)} = c_\mu \norm{\frac{\Lambda}{c_\mu} \, \theta^*}_{M_t^{-1}(c_\mu)} \leq \sqrt{c_\mu} \norm{\theta^*}_{\Lambda} \leq \sqrt{c_\mu}(\sqrt{\lambda_0}S_0 + \sqrt{\lambda_\perp} S_\perp). \label{prop_pf_3}
\end{align}

Combine Equation \eqref{prop_pf_2} and \eqref{prop_pf_3} into Equation \eqref{prop_pf_1}, we finish our proof. \hfill \qedsymbol
\par Since Equation \eqref{prop_mubound_equ} in Proposition \ref{prop_mubound} holds simultaneously for all $x \in \R$ and $t \geq 1$, the following conclusion holds.
\begin{corollary} \label{coro_rvbound}
For any random variable $z$ defined in $\R$, we have the following holds
\begin{align}
     \vert \mu(z^\top \theta^*) - \mu(z^\top \hat\theta_t)\vert \leq \beta_{t+T_1}^z(\delta), \nonumber
\end{align}
with probability at least $1-\delta$. Furthermore, for any sequence of random variable $\{z_t\}_{t=2}^{T}$, with probability $1-\delta$ it holds that
\begin{align}
     \vert \mu(z_t^\top \theta^*) - \mu(z_t^\top \hat\theta_t)\vert \leq \beta_{t+T_1}^{z_t}(\delta), \nonumber
\end{align}
simultaneously for all $t \geq 1$.
\end{corollary}

\subsubsection{Proof of Lemma \ref{lemma_goodprob}}\label{subsec:lemma_goodprob}
For the proof of Lemma \ref{lemma_goodprob} we will need the following two lemmas, and we will use the same notations as in Lemma \ref{lemma_goodprob} in this section.
\begin{lemma}\label{lem:goodprob_lem1}
Let $\lambda \in \R^d$ be arbitrary and consider any $t \geq 0$
$$M_t^\lambda = \exp \left( \sum_{s=1}^t  \left[ \frac{\eta_s(\lambda^\top x_s)}{\sigma_0} - \frac{1}{2} (\lambda^\top x_s)^2 \right]\right).$$
Let $\tau$ be a stopping time with respect to the filtration $\{\mathcal{F}_t\}_{t=0}^{+\infty}$. Then $M_t^\lambda$ is a.s. well defined and $\E(M_\tau^\lambda) \leq 1$.
\end{lemma}
\proof We claim that $\{M_t^\lambda\}$ is a supermartingale. Let

$$D_t^{\lambda} = \exp \left( \frac{\eta_s(\lambda^\top x_s)}{\sigma_0} - \frac{1}{2} (\lambda^\top x_s)^2 \right)$$

Observe that by conditional $\sigma_0$-sub-Gaussianity of $\eta_t$ we have $\E[D_t^{\lambda} | \mathcal{F}_{t-1}] \leq 1$. Clearly, $D_t^{\lambda}$ and $M_t^{\lambda}$ is $\mathcal{F}_{t}$-measurable. Moreover,
$$\E[M_t^{\lambda} | \mathcal{F}_{t-1}] = \E[M_{1}^{\lambda} \cdots D_{t-1}^{\lambda} D_t^{\lambda}| \mathcal{F}_{t-1}] = D_1^{\lambda} \dots D_{t-1}^{\lambda} \E[D_t^{\lambda} | \mathcal{F}_{t-1}] \leq M_{t-1}^{\lambda},$$
which implies that $M_{t}^{\lambda}$ is a supermartingale with its expected value upped bounded by $1$. To show that $M_{t}^{\lambda}$ is well defined. By the convergence theorem for nonnegative supermartingales, $\lim_{t \rightarrow \infty} M_{t}^{\lambda}$ is a.s. well-defined, which indicates that $M_{\tau}^{\lambda}$ is also well-defined for all $\tau \in \mathbb{N}^+ \cup \{+\infty\}$. By Fatou's Lemma, it holds that
$$\E[M_{\tau}^\lambda] = \E[\lim \inf_{t \rightarrow \infty} M_{\min \{t,\tau\}}^\lambda] \leq \lim \inf_{t \rightarrow \infty} \E[ M_{\min\{t,\tau\}}^\lambda] \leq 1.$$
\hfill \qedsymbol
\begin{lemma} \label{lem:goodprob_lem2}
For any positive semi-definite matrix $P \in \R^{d \times d}$ and positive definite matrix $Q \in \R^{d \times d}$, and any $x,a \in \R^d$, it holds that
$$\norm{x-a}_P^2 + \norm{x}_Q^2 = \norm{x - (P+Q)^{-1}Pa}_{P+Q}^2 + \norm{a}_P^2 - \norm{Pa}_{(P+Q)^{-1}}^2.$$
\end{lemma}
This lemma could be easily proved based on elementary calculation and hence its proof would be omitted here.
\begin{lemma}\label{lem:goodprob_lem3}
Let $\tau$ be a stopping time with $\tau > m$ on the filtration $\{\mathcal{F}_t\}_{t=0}^{\infty}$. Then for $\delta > 0$, with probability $1-\delta$,
$$||S_\tau||_{V_\tau^{-1}}^2 \leq 2 \sigma_0^2 \log \left(\frac{\text{det}(V_\tau)^{1/2} \text{det}(\Lambda)^{-1/2}}{\delta} \right).$$
\end{lemma}
\proof W.l.o.g., assume that $\sigma_0 = 1$. Denote
$$\tilde V_t = V_t-\Lambda = \sum_{s=1}^t x_sx_s^\top, \quad \qquad M_{t}^\lambda = \exp \left( (\lambda^\top S_t) - \frac{1}{2} ||\lambda||_{\tilde V_t}^2 \right).$$
Note by Lemma \ref{lem:goodprob_lem1}, we naturally have that $\E[M_{t}^\lambda] \leq 1$.

Since in round $m+1$, we get the diagonal positive definite matrix $\Lambda$ with its elements independent with samples after round $m$. Let $z$ be a Gaussian random variable that is independent with other random variables after round $m$ with covariance $\Lambda^{-1}$. Define
$$M_t = \E[M_t^{z} | \mathcal{F}_{\infty}], \quad t>m,$$
where $\mathcal{F}_{\infty}$ is the tail $\sigma$-algebra of the filtration. Clearly, it holds that $\E[M_\tau] = \E[\E[M_\tau^z | z,\mathcal{F}_{\infty}]] \leq \E[1] \leq 1$. Let $f$ be the density of $z$ and for a positive definite matrix $P$ let $c(P) = \sqrt{(2\pi)^d/\text{det}(P)}$. Then for $t>m$ it holds that,
\begin{align*}
    M_t &= \int_{\R^d} \exp \left( (\lambda^\top S_t) - \frac{1}{2} ||\lambda||_{\tilde V_t}^2 \right) f(\lambda) d \lambda \\
    &= \frac{1}{c(\Lambda)} \exp\left(\frac{1}{2} \norm{S_t}_{\tilde V_t^{-1}}^2\right) \int_{\R^d} \exp\left( -\frac{1}{2} \left\{\norm{\lambda - \tilde V_t^{-1} S_t}_{\tilde V_t}^2 + \norm{\lambda}_\Lambda^2 \right\} \right) d \lambda.
\end{align*}
Based on Lemma \ref{lem:goodprob_lem2}, it holds that
$$\norm{\lambda - \tilde V_t^{-1} S_t}_{\tilde V_t}^2 + \norm{\lambda}_\Lambda^2 = \norm{\lambda - V_t^{-1} S_t}_{V_t}^2 + \norm{\tilde V_t^{-1} S_t}_{\tilde V_t}^2 - \norm{S_t}_{V_t^{-1}}^2$$,
and this implies that
\begin{align*}
    M_t &= \frac{1}{c(\Lambda)} \exp\left( \frac{1}{2} \norm{S_t}_{V_t^{-1}}^2 \right) \int_{\R^d} \exp \left( -\frac{1}{2} \norm{\lambda - V_t^{-1} S_t}_{V_t}^2 \right) d\lambda \\
    &= \left( \frac{\text{det}(\Lambda)}{\text{det}(V_t)} \right)^{1/2}  \exp \left( -\frac{1}{2} \norm{\lambda - V_t^{-1} S_t}_{V_t}^2 \right).
\end{align*}
Now, from $\E[M_\tau] \leq 1$, we have that for $\tau > m$
\begin{align*}
    P\left( \norm{S_\tau}_{V_\tau^{-1}}^2 >  \log \left(\frac{\text{det}(V_\tau)}{\delta^2  \text{det}(\Lambda)} \right) \right) &= P \left( \frac{\exp \left(\frac{1}{2} \norm{S_\tau}_{V_\tau^{-1}}^2 \right)}{\delta^{-1} (\text{det}(V_\tau)/\text{det}(\Lambda))^{1/2} }  > 1 \right) \\
    &\leq \E \left[ \frac{\exp \left(\frac{1}{2} \norm{S_\tau}_{V_\tau^{-1}}^2 \right)}{\delta^{-1} (\text{det}(V_\tau)/\text{det}(\Lambda))^{1/2} } \right] \leq \E[M_\tau] \delta \leq \delta.
\end{align*}
\hfill \qedsymbol

Combining Lemma \ref{lem:goodprob_lem1}-\ref{lem:goodprob_lem3}. We now contruct a stopping time and define the bad event:
$$B_t(\delta) \coloneqq \left\{ w : \norm{S_t}_{V_t^{-1}}^2 > \sigma_0^2 \log \left( \frac{\text{det}(V_t)}{\delta^2  \text{det}(\Lambda)} \right) \right\}.$$
And we are interested in bounding the probability that $\cup_{t>m} B_t(\delta)$ happens. Define $\tau(w) = \min \{t >m: w \in B_t(\delta)\}$. Then $\tau$ is a stopping time and it holds that,
$$\cup_{t >m}B_t(\delta) = \{w : \tau(w) < \infty\}.$$
Then we have that
\begin{align*}
    P\left[ \cup_{t >m}B_t(\delta) \right] = P [m<\tau < \infty] = P\left[\norm{S_\tau}_{V_\tau^{-1}}^2 > \sigma_0^2 \log \left( \frac{\text{det}(V_\tau)}{\delta^2  \text{det}(\Lambda)} \right), \tau > m\right] \leq \delta.
\end{align*}
This concludes our proof of Lemma \ref{lemma_goodprob}.

\subsection{Proposition \ref{prop_mubound2} with its proof}
We denote the optimal action $x^* = \arg \max_{x \in \mathcal{X}_0} \mu(x^\top \theta^*)$. 
\begin{proposition} \label{prop_mubound2}
For all $\delta \in (0,1)$, with probability $1 - \delta$, it holds that
\begin{align}
    \mu({x^*}^\top \theta^*) - \mu(x_t^\top \theta^*) \leq 2 \beta_{t+T_1}^{x_t}\left(\frac{\delta}{2}\right), \nonumber
\end{align}
simultaneously for all $t \in \{2,3,\dots,T_2\}$.
\end{proposition}
\proof According to Corollary \ref{coro_rvbound}, outside of the event of measure can be bounded by $\delta/2$:
\begin{align}
    \mu(x_t^\top \hat\theta_t) - \mu(x_t^\top \theta^*) \leq \beta_{t+T_1}^{x_t}\left(\frac{\delta}{2}\right) \quad \text{for all } t \in \{2,3,\dots,T_2\}. \nonumber
\end{align}
Similarly, with probability at least $1-\delta/2$ it holds that
\begin{align}
    \mu(x^*{}^\top \theta^*) - \mu(x^*{}^\top \hat\theta_t) \leq \beta_{t+T_1}^{x^*}\left(\frac{\delta}{2}\right) \quad \text{for all } t \in \{2,3,\dots,T_2\}. \nonumber
\end{align}
Besides, by the choice of $x_t$ in Algorithm \ref{alg1}
\begin{align}
    \mu(x^*{}^\top \hat\theta_t) - \mu(x_t^\top \hat\theta_t) &= \mu(x^*{}^\top \hat\theta_t) + \beta_{t+T_1}^{x^*}\left(\frac{\delta}{2}\right) - \mu(x_t^\top \hat\theta_t) - \beta_{t+T_1}^{x^*}\left(\frac{\delta}{2}\right) \nonumber \\ 
    & \leq  \mu(x_t^\top \hat\theta_t) + \beta_{t+T_1}^{x_t}\left(\frac{\delta}{2}\right) - \mu(x_t^\top \hat\theta_t) - \beta_{t+T_1}^{x^*}\left(\frac{\delta}{2}\right) \nonumber \\ &= \beta_{t+T_1}^{x_t}\left(\frac{\delta}{2}\right) - \beta_{t+T_1}^{x^*}\left(\frac{\delta}{2}\right). \nonumber
\end{align}

By combining the former inequalities we finish our proof. \hfill \qedsymbol

\subsection{Proof of Theorem \ref{thm_regalg1}}
\proof Based on Proposition \ref{prop_mubound2} we have
\begin{align}
    \mu({x^*}^\top \theta^*) - \mu(x_t^\top \theta^*) \leq 2 \beta_{{t+T_1}}^{x_t}\left(\frac{\delta}{2}\right)
 = 2 \alpha_{{t+T_1}}\left(\frac{\delta}{2}\right)
 \norm{x_t}_{M_t^{-1}(c_\mu)} \leq 2 \alpha_{T}\left(\frac{\delta}{2}\right)
 \norm{x_t}_{M_t^{-1}(c_\mu)}. \nonumber
\end{align}

Since we know that $\mu({x^*}^\top \theta^*) - \mu(x^\top \theta^*) \leq k_\mu({x^*}^\top \theta^* - x^\top \theta^*) \leq 2 k_\mu S_0^2$ for all possible action $x$, and we can safely expect that $\alpha_{T_2}(\delta/2) > k_\mu S_0^2$ (at least by choosing $\sigma_0 = k_\mu \max\{S_0^2,1\}$), then the regret of Algorithm \ref{alg1} can be bounded as
\begin{align}
    Regret_{T_2} &\leq 2 k_\mu S_0^2 + \sum_{t=2}^{T_2} \min\{\mu({x^*}^\top \theta^*) - \mu(x_t^\top \theta^*), 2 k_\mu S_0^2\} \nonumber \\ 
    &\leq 2 k_\mu S_0^2 + 2 \alpha_{T}\left(\frac{\delta}{2}\right) \sum_{t=2}^{T_2} \min\{\norm{x_t}_{M_t^{-1}(c_\mu)}, 1\} \nonumber \\ 
    &\stackrel{\textup{(i)}}{\leq} 2 k_\mu S_0^2 + 2 \alpha_{T}\left(\frac{\delta}{2}\right)\sqrt{T_2} \sqrt{\sum_{t=2}^{T_2} \min\{\norm{x_t}_{M_t^{-1}(c_\mu)}^2, 1\}}. \nonumber
\end{align}
where the ineuqlity (i) comes from Cauchy-Schwarz inequality. And a commonly-used fact (e.g.~\cite{abbasi2011improved}, Lemma 11) yields that
\begin{align}
    \sum_{i=2}^{t}\min\{\norm{x_i}_{M_i^{-1}(c_\mu)}^2, 1\} &\leq 2 \log\left( \dfrac{\vert M_{t+1}(c_\mu)\vert}{\vert M_{2}(c_\mu) \vert} \right) \leq  2\log\left( \dfrac{\vert M_{t+1}(c_\mu)\vert}{\vert \frac{\Lambda}{c_\mu} \vert} \right) \nonumber \\
    &\leq  k \log \left(1+\frac{c_\mu S_0^2}{k \lambda_0}(t+T_1) \right) +  \frac{c_\mu S_0^2}{\lambda_\perp}(t+T_1). \nonumber
\end{align}

Finally, by using the argument in Eqn. \eqref{prop_pf_2} and then plugging in the chosen value for $\lambda_\perp = \dfrac{c_\mu S_0^2 T}{k \log(1+\frac{c_\mu S_0^2 T}{k \lambda_0})}$, we have  
\begin{align}
    &Regret_{T_2} \leq 2 k_\mu S_0^2 + \nonumber \\ 
    &\frac{2k_\mu}{c_\mu} \left(\sigma_0 \sqrt{2k \log \left(1 + \frac{c_\mu S_0^2}{k \lambda_0}T \right) - 2\log \left(\frac{\delta}{2}\right)} +  \sqrt{c_\mu}\left(\sqrt{\lambda_0} S_0 + \sqrt{\frac{c_\mu S_0^2 T}{k\log \left(1 + \frac{c_\mu S_0^2}{k \lambda_0}T \right)}}S_\perp\right)\right) \nonumber \\
    &\times \sqrt{T_2} \sqrt{4k\log \left(1 + \frac{c_\mu S_0^2}{k \lambda_0}T \right)}, \nonumber
\end{align}
which gives us the final bound in Theorem \ref{thm_regalg1}. \hfill \qedsymbol

\section{Consistency of $\hat\theta_t^{\new}$ in Algorithm \ref{alg1}}\label{sec:app:ucbglm}
W.l.o.g. we assume that $\{\theta \, : \, \norm{\theta-\theta^*}_2 \leq 1\} 	\subseteq \mathit{\Theta}^*$, or otherwise we can modify the contraint of $c_\mu$ in Assumption \ref{assu_link} as $c_\mu \coloneqq \inf_{\{x \in \mathcal{X}_0, \norm{\theta-\theta^*}_2 \leq 1\}} \mu^{\prime}(x^\top \theta) > 0$. And we also assume that $\norm{x}_2 \leq 1$ for $x \in \mathcal{X}_0$. 


\par Adapted from the proof of Theorem 1 in~\cite{li2017provably}, define $G(\theta) = g(\theta) - g(\theta^*)=\sum_{i=1}^{T_1} (\mu(x_{s_1,i}^\top \theta) - \mu({s_1,i}^\top \theta^*))x_{s_1,i}+\sum_{i=1}^n (\mu(x_i^\top \theta) - \mu(x_i^\top \theta^*))x_i + \Lambda(\theta - \theta^*)$.
W.l.o.g we suppose $c_\mu \leq 1$ based on argument in Appendix \ref{sec:app:remark}. Then it holds that for any $\theta_1,\theta_2 \in \R^p$
\begin{align*}
    &G(\theta_1) - G(\theta_2) = \\ &\left[ \sum_{i=1}^{T_1} (\mu^\prime(x_{s_1,i}^\top \theta) - \mu({s_1,i}^\top \theta^*))x_{s_1,i} x_{s_1,i}^\top+\sum_{i=1}^n (\mu^\prime(x_i^\top \theta) - \mu(x_i^\top \theta^*))x_i x_i^\top + \Lambda \right] (\theta_1 - \theta_2).
\end{align*}

By denoting $V = \sum_{i=1}^{T_1} x_{s_1,i} x_{s_1,i}^\top +  \sum_{i=1}^n  x_i x_i^\top + \Lambda$. We have
$$(\theta_1 - \theta_2)^\top (G(\theta_1) - G(\theta_2)) \geq (\theta_1 - \theta_2)^\top (c_\mu V) (\theta_1 - \theta_2)>0$$
Therefore, the rest of proof would be identical to that of Step 1 in the proof of Theorem 1 in \cite{li2017provably}.
Based on the step 1 in the proof of Theorem 1 in~\cite{li2017provably}, we have
\begin{align*}
    \norm{G(\theta)}_{V^{-1}}^2 \geq c_\mu^2 \lambda_{\min}(V) \norm{\theta-\theta^*}_2^2.
\end{align*}
as long as $\norm{\theta-\theta^*}_2 \leq 1$. Then Lemma A of~\cite{chen1999strong} and Lemma 7 of~\cite{li2017provably} suggest that we have 
\begin{align*}
    \norm{\hat\theta - \theta^*} \leq \frac{4\sigma}{c_\mu} \sqrt{\frac{p+\log(1/\delta)}{\sigma^2}} \leq 1,
\end{align*}
when $\lambda_{\min}(V) \geq 16 \sigma^2 [p+\log(1/\delta)]/c_\mu^2$ for any $\delta > 0$. Therefore, it suffices to show that the condition $\lambda_1 \geq 16 \sigma^2 [p+\log(1/\delta)]/c_\mu^2$ for any $\delta > 0$ holds with high probability (e.g. $1-\delta$), and we utilize the Proposition 1 of~\cite{li2017provably}, which is given as follows:
\vspace{0.1 cm}
\newline \textbf{Proposition} (Proposition 1 of~\cite{li2017provably}): \textit{Define $V_n = \sum_{t=1}^n x_tx_t^\top (+\Lambda)$ where $x_i$ is drawn iid from some distribution $\nu$ with suppost in the unit ball, $\mathbb{B}^d$. Furthermore, let $\Sigma = \mathbb{E}(x_tx_t^\top)$ be the second moment matrix, and $B$ and $\delta$ be two positive constants. Then, there exists positive universal constants $C_1$ and $C_2$ such that $\lambda_{\min}(V_n) \geq B$ with probability at least $1-\delta$, as long as
\begin{align*}
    n \geq \left(\frac{C_1 \sqrt{d} + C_2 \sqrt{\log(1/\delta)}}{\lambda_{\min}(\Sigma)} \right)^2 + \frac{2B}{\lambda_{\min}(\Sigma)}
\end{align*}}

Therefore, we can dedeuce that $\norm{\hat \theta_t - \theta^*}_2 \leq 1$ holds with probability at least $1-\delta$ as long as $T_1 \geq ((\hat{C}_1\sqrt{p}+\hat{C}_2\sqrt{\log(1/\delta))}/{\lambda_1})^2 + {2 B}/{\lambda_1}$ holds for some absolute constants $\hat{C}_1,\hat{C}_2$ with the definition $B \coloneqq 16\sigma^2(p+\log(1/\delta))/c_\mu^2$. Notice that this condition could easily hold if $\lambda_1 \asymp \sigma^2$ is not diminutive in magnitude. Otherwise, we believe a tighter bound exists in that case, and we will leave it as a future work.

We also present an intuitive explanation for this consistency result: 
\cite{li2017provably} proved the consistency of the MLE $\hat\theta_{t}$ without the regularizer. Regarding the penalty $\theta^\top \Lambda \theta$, for the first $k$ entries of $\hat\theta_{t}$ the penalized parameter $\lambda_0$ is small, and hence it will have mild effect after sufficient warm-up rounds $T_1$. For the remaining $(p-k)$ elements suffering large penalty, the estimated $\hat \theta_{t,k+1:p}$ would be ultra small in magnitude as desired since we argue that after the transformation ${\theta_{k+1:p}^*}$ will also be insignificant.
This implies that $\norm{\hat \theta_{t,k+1:p}-\theta_{k+1:p}^*}_2$ is well contronlled. 
As a result, the estimated $\hat \theta_t$ tends to be consistent.
\section{Analysis of Theorem \ref{thm_regestt}}

\subsection{Proof of Theorem \ref{thm_regestt}}
\proof Let us define $r_t = \max_{X \in \mathcal{X}} \mu(\inp{X}{\Theta^*}) - \mu(\inp{X_t}{\Theta^*})$, the instantaneous regret at time $t$. We can easily bound the regret for stage 1 as $\sum_{t=1}^{T_1} r_t \leq 2 S_f T_1$. For the second stage, we have a bound according to Theorem \ref{thm_regalg1} (Theorem \ref{thm_regalg2}):
\begin{align}
    \sum_{t=T_1+1}^{T} r_t \leq \widetilde{O}(k \sqrt{T} + \sqrt{ \lambda_0 k T} \, + T S_\perp)
    \leq \widetilde{O} \left(k \sqrt{T} + \sqrt{ \lambda_0 k T} \, + T \frac{(d_1+d_2)Mr}{T_1 D_{rr}^2}\log\left(\frac{d_1+d_2}{\delta}\right) \right). \nonumber
\end{align}

Therefore, the overall regret is:
\begin{align}
    \sum_{t=1}^{T} r_t \leq \widetilde{O} \left(2S_f  T_1 + k \sqrt{T} + \sqrt{ \lambda_0 k T} \, + T \frac{(d_1+d_2)Mr}{T_1 D_{rr}^2}\log\left(\frac{d_1+d_2}{\delta}\right) \right). \nonumber
\end{align}

After plugging the choice of $T_1$ given in Theorem \ref{thm_regestt}, it holds that
\begin{align}
    \sum_{t=1}^{T} r_t &\leq \widetilde{O} \left((\frac{\sqrt{r(d_1+d_2)M}}{D_{rr}} + \sqrt{\lambda_0 k} + k)\sqrt{T} \right) \lesssim \widetilde{O} \left((\frac{\sqrt{r(d_1+d_2)M}}{D_{rr}} + k) \sqrt{T} \right) \nonumber \\
    &= \widetilde{O} \left(\frac{\sqrt{(d_1+d_2)MrT}}{D_{rr}} \right). \nonumber
\end{align}  



\hfill \qedsymbol

\section{Details of Theorem \ref{thm_regests}}
\subsection{Proof of Theorem \ref{thm_regests}}
\proof Here we will overload the notation a little bit. Under the new arm feature set and parameter set after rotation, let $X^*$ be the best arm and $X_t$ be the arm we pull at round $t$ for stage 2. And we denote $x_{t,sub}$ be the vectorization of $X_t$ after removing the last $p-k$ covariates, and similarly define $x^*_{sub}$ and $\theta^*_{sub}$ as the subtracted version of $\vecc{X^*}$ and $\vecc{\Theta^*}$ respectively. We use $r_t = \mu(\inp{X^*}{\Theta^*}) - \mu(\inp{X_t}{\Theta^*})$ as the instantaneous regret at round t for stage 2. Then it holds that, for $t \in [T_2]$
\begin{align}
    r_t &= \mu(\inp{X^*}{\Theta^*}) - \mu(x^*_{sub}{}^\top \theta^*_{sub}) + 
    \mu(x^*_{sub}{}^\top \theta^*_{sub}) - \mu(x_{t,sub}^\top \theta^*_{sub}) + \mu(x_{t,sub}^\top \theta^*_{sub}) - \mu(\inp{X_t}{\Theta^*}) \nonumber \\
    &\leq k_\mu \vert \inp{X^*}{\Theta^*} - x^*_{sub}{}^\top \theta^*_{sub} \vert + k_\mu \vert \inp{X_t}{\Theta^*} - x_{t,sub}^\top \theta^*_{sub} \vert + \mu(x^*_{sub}{}^\top \theta^*_{sub}) - \mu(x_{t,sub}^\top \theta^*_{sub}) \nonumber \\ 
    &\leq k_\mu(\norm{\widehat U_\perp^\top X^* \widehat V_\perp}_F+\norm{\widehat U_\perp^\top X_t \widehat V_\perp}_F) \norm{\widehat U_\perp^\top U D V^\top \widehat V_\perp}_F +  \mu(x^*_{sub}{}^\top \theta^*_{sub}) - \mu(x_{t,sub}^\top \theta^*_{sub}) \nonumber \\ 
    &\leq 2 k_\mu S_0 \dfrac{d_1d_2r}{T_1 D_{rr}^2}\log\left(\frac{d_1+d_2}{\delta}\right) +\mu(x^*_{sub}{}^\top \theta^*_{sub}) - \mu(x_{t,sub}^\top \theta^*_{sub}). \nonumber
\end{align}

Therefore, the overall regret can be bounded as
\begin{align}
    2 S_f  T_1+ \sum_{t=1}^{T_2} r_t \leq 2 S_f  T_1 + 2 k_\mu S_0 \frac{d_1d_2r}{D_{rr}^2 T_1 }T_2 + \sum_{t=1}^{T_2} \mu(x^*_{sub}{}^\top \theta^*_{sub}) - \mu(x_{t,sub}^\top \theta^*_{sub}). \nonumber
\end{align}

Since efficient low dimensional generalized linear bandit algorithm can achieve regret $\widetilde{O}(\epsilon \sqrt{d} T)$ where $\epsilon$ is the misspecified rate, $d$ is the dimension of parameter and $T$ is the time horizon when no sparsity (low-rank structure) presents in the model. After plugging our carefully chosen $T_1$, the regret is 
\begin{align}
    2 S_f T_1 &+ 2 k_\mu S_0 \dfrac{(d_1+d_2)Mr}{T_1 D_{rr}^2}\log\left(\frac{d_1+d_2}{\delta}\right)T_2 + \widetilde{O}  \left( \dfrac{(d_1+d_2)Mr}{T_1 D_{rr}^2} \sqrt{(d_1+d_2)r} T_2 \right) \nonumber \\
    &= \widetilde{O} \left((\frac{\sqrt{r^{3/2}(d_1+d_2)^{3/2}M}}{D_{rr}} + k) \sqrt{T} \right) = \widetilde{O}\left(\frac{\sqrt{r^{3/2}(d_1+d_2)^{3/2}M}}{D_{rr}}\sqrt{T}\right). \label{eq:ests}
\end{align} 
\hfill \qedsymbol
%
\section{Explanation of $V_t$ replacing $M_t(c_\mu)$} \label{sec:app:remark}
Technically we can always assume $c_\mu \in (0,1]$ since we can always choose $c_\mu = 1$ when it can take values greater than $1$. And when $c_\mu \leq 1$ it holds that,
\begin{align}
    M_t(c_\mu) = \sum_{i=1}^{t-1} x_i x_i^\top + \dfrac{\Lambda}{c_\mu} \succeq \sum_{i=1}^{t-1} x_i x_i^\top + \Lambda = V_t. \nonumber
\end{align}
Therefore, we can easily keep the exactly identical outline of our proof of the bound of regret for Algorithm \ref{alg1} after replacing $M_t(c_\mu)$ by $V_t$ everywhere, and the result only change by a constant factor of $1/\sqrt{c_\mu}$, which would not be too large in most cases. However, in our algorithm and proof we still use $M_t(c_\mu)$ for a better theoretical bound.

\section{Additional Algorithms}
\subsection{PLowGLM-UCB}\label{app:plow}
We could modify Algorithm \ref{alg1} by only recomputing $\hat\theta_t$ and whenever $\vert M_t(c_\mu)\vert$ increases by a constant factor $C > 1$ in scale, and consequently we only need to solve the Eqn. \eqref{hattheta} for $O(\log(T_2))$ times up to the horizon $T_2$, which significantly alleviate the computational complexity. The pseudo-code of PLowGLM-UCB is given in Algorithm \ref{alg2}.

\begin{algorithm}
\caption{PLowGLM-UCB} \label{alg2}
\begin{algorithmic}[1] 
\Input $T_2,k,\mathcal{X}_0$, the probability rate $\delta$, penalization parameters $(\lambda_0,\lambda_\perp)$, the constant $C$.
\State Initialize $ M_1(c_\mu) = \sum_{i=1}^{T_1} x_{s_1,i} \, x_{s_1,i}^\top + \Lambda/c_\mu$. 
\For{$t \geq 1$}
\If{$\vert M_t(c_\mu)\vert > C \vert M_\tau (c_\mu)\vert$}
\State Estimate $\hat \theta_t$ according to \eqref{hattheta}.
\State $\tau = t$
\EndIf
\State Choose arm $x_t = \arg\max_{x \in \mathcal{X}_0} \{\mu({x}^\top{\btheta_\tau}) + \rho_{\tau}(\delta)\norm{x}_{M_t^{-1}(c_\mu)} \}$, receive $y_t$.
\State Update $M_{t+1}(c_\mu) \longleftarrow M_{t}(c_\mu) + x_t x_t^\top$.
\EndFor
\end{algorithmic}
\end{algorithm}
Theorem \ref{thm_regalg2} shows the regret bound of PLowGLM-UCB under Assumption \ref{assu_bound} and \ref{assu_link}.
\begin{theorem} \label{thm_regalg2} \textup{(Regret of PLowGLM-UCB)}
For any fixed failure rate $\delta \in (0,1)$, if we run the PLowGLM-UCB algorithm with $\rho_t(\delta) = \alpha_{t+T_1}(\delta/2)$ and
\begin{align}
 \lambda_\perp \asymp \dfrac{c_\mu S_0^2 T}{k \log(1+\frac{c_\mu S_0^2 T}{k \lambda_0})}. \nonumber
\end{align}
Then the regret of PLowGLM-UCB $(Regret_{T_2})$ satisfies, with probability at least $1-\delta$
\begin{align}
    \widetilde{O}(k \sqrt{T_2} + \sqrt{ \lambda_0 k T} \, + T S_\perp) \cdot \sqrt{C} {= \widetilde{O}(k \sqrt{T}+ T S_\perp) \cdot \sqrt{C}}. \nonumber
\end{align}
\end{theorem}
Similarly, for PLowUCB-GLM we can also prove that the regret bound increase at most by a constant multiplier $\sqrt{C}$ by using the same lemma and argument we show in the following Section \ref{app:thm_regalg2}. And we can get the bound of regret for PLowGLM-UCB in problem dependence case, and the bound will be exactly the same as that we have shown in Theorem \ref{thm_regalg1} except a constant multiplier $\sqrt{C}$. 

\subsection{Proof of Theorem \ref{thm_regalg2}}\label{app:thm_regalg2}
We use similar sketch of proof for Theorem 5 in~\cite{abbasi2011improved}. First, we show the following lemma:
\begin{lemma} \label{lemma_twopd} \textup{(\cite{abbasi2011improved}, Lemma 12)}
Let $A$ and $B$ be two positive semi-definite matrices such that $A \preceq B$. Then, we have that
\begin{align}
    \sup_{x \neq 0} \dfrac{x^\top A x}{x^\top B x} \leq \dfrac{\vert A \vert}{\vert B \vert}. \nonumber
\end{align}
\end{lemma}
Then we can outline the proof of Theorem \ref{thm_regalg2} as follows.
\proof Let $\tau_t$ be the value of $\tau$ at step $t$ in Algorithm \ref{thm_regalg2}. By an argument similar to the one used in proof of Theorem \ref{thm_regalg1}, we deduce that for any $x \in \R$ and all $t \geq 2$ simultaneously,
\begin{align}
    \vert \mu(x^\top \theta^*) - \mu(x^\top \hat\theta_{\tau_t}) \vert &\leq \frac{k_\mu}{c_\mu} \norm{g_{\tau_t}(\theta^*) - g_{\tau_t}(\hat\theta_{\tau_t})}_{M_{\tau_t}^{-1}(c_\mu)} \norm{x}_{M_{\tau_t}^{-1}(c_\mu)}  \nonumber \\
    &= \frac{k_\mu}{c_\mu} \norm{g_{\tau_t}(\theta^*) - g_{\tau_t}(\hat\theta_{\tau_t})}_{M_{\tau_t}^{-1}(c_\mu)} \norm{{M_{\tau_t}^{-\frac{1}{2}}}(c_\mu)\,x}_2 \nonumber \\
    &\leq \frac{k_\mu}{c_\mu} \norm{g_{\tau_t}(\theta^*) - g_{\tau_t}(\hat\theta_{\tau_t})}_{M_{\tau_t}^{-1}(c_\mu)} \norm{{M_{t}^{-\frac{1}{2}}}(c_\mu)\,x}_2 \sqrt{\frac{\vert M_{\tau_t}^{-1}(c_\mu) \vert}{\vert M_{t}^{-1}(c_\mu) \vert}} \nonumber \\
    &\leq \frac{k_\mu}{c_\mu}  \sqrt{C} \norm{g_{\tau_t}(\theta^*) - g_{\tau_t}(\hat\theta_{\tau_t})}_{M_{\tau_t}^{-1}(c_\mu)} \norm{x}_{M_{t}^{-1}(c_\mu)} \leq \sqrt{C} \beta_{t+T_1}^x(\delta). \nonumber
\end{align}
where the last inequality comes from the proof of Proposition \ref{prop_mubound} similarly. The rest of the proof will be mostly identical to that of Theorem \ref{thm_regalg1} and hence we would copy it here for completeness:
\par Based on Proposition \ref{prop_mubound2} we have
\begin{align}
    \mu({x^*}^\top \theta^*) - \mu(x_t^\top \theta^*) \leq 2\sqrt{C} \beta_{t+T_1}^{x_t}\left(\frac{\delta}{2}\right)
 &= 2\sqrt{C} \alpha_{t+T_1}\left(\frac{\delta}{2}\right) 
 \norm{x_t}_{M_t^{-1}(c_\mu)} \nonumber  \\
 &\leq 2\sqrt{C} \alpha_{T}\left(\frac{\delta}{2}\right)
 \norm{x_t}_{M_t^{-1}(c_\mu)}. \nonumber
\end{align}

Since we have that $ \alpha_{T_2}(\delta/2) >  k_\mu S_0^2$, the Regret of Algorithm \ref{alg2} can be bounded as
\begin{align}
    Regret_{T_2} &\leq 2 k_\mu S_0^2 + \sum_{t=2}^{T_2} \min\{\mu({x^*}^\top \theta^*) - \mu(x_t^\top \theta^*), 2 k_\mu S_0^2\} \nonumber \\ 
    &\leq 2 k_\mu S_0^2 + 2\sqrt{C} \alpha_{T}\left(\frac{\delta}{2}\right) \sum_{t=2}^{T_2} \min\{\norm{x_t}_{M_t^{-1}(c_\mu)}, 1\} \nonumber \\ 
    &\leq 2 k_\mu S_0^2 + 2\sqrt{C} \alpha_{T}\left(\frac{\delta}{2}\right)\sqrt{T_2} \sqrt{\sum_{t=2}^{T_2} \min\{\norm{x_t}_{M_t^{-1}(c_\mu)}^2, 1\}}. \nonumber
\end{align}
where the last ineuqlity comes from Cauchy-Schwarz inequality. Finally, by a self-normalized martingale inequality (~\cite{abbasi2011improved}, Lemma 11) and and then plugging in the chosen value for $\lambda_\perp = \dfrac{c_\mu S_0^2 T}{k \log(1+\frac{c_\mu S_0^2 T}{k \lambda_0})}$, we have  
\begin{align}
    &Regret_{T_2} \leq 2 k_\mu S_0^2 + \frac{2k_\mu}{c_\mu}\sqrt{C} \nonumber \\
    & \times \left(\sigma_0 \sqrt{2k \log \left(1 + \frac{c_\mu S_0^2}{k \lambda_0}T \right) - 2\log \left(\frac{\delta}{2}\right)} + \sqrt{c_\mu}\left(\sqrt{\lambda_0} S_0 + \sqrt{\frac{c_\mu S_0^2 T}{k\log \left(1 + \frac{c_\mu S_0^2}{k \lambda_0}T \right)}}S_\perp\right)\right) \nonumber \\
    & \times \sqrt{T_2} \sqrt{4k\log \left(1 + \frac{c_\mu S_0^2}{k \lambda_0}T \right)}, \nonumber
\end{align}
which gives us the final bound in Theorem \ref{thm_regalg2}. \hfill \qedsymbol


\begin{algorithm}[t]
\caption{Generalized Explore Subspace Then Transform (G-ESTT)} \label{alg_estt_con}
\begin{algorithmic}[1]
\Input Action set $\{\mathcal{X}_t\}$, $T,T_1,\mathcal{D}$, the probability rate $\delta$, parameters for stage 2: $\lambda, \lambda_\perp$. \vspace{0.07 cm}
\Stage \textbf{1: Subspace Estimation}

\State Randomly choose $X_t \in \mathcal{X}$ according to $\mathcal{D}$ and record $X_t, Y_t$ for $t = 1,\dots T_1$.
\State Obtain $\widehat \Theta$ by solving the following equation:
\begin{align}
    \widehat \Theta = \arg \min_{\Theta \in R^{d_1 \times d_2}} {\dfrac{1}{T_1} \sum_{i = 1}^{T_1} \{b(\inp{X_i}{\Theta}) - y_i \inp{X_i}{\Theta}\}} + \lambda_{T_1} \nunorm{\Theta}.  \nonumber 
\end{align}
\State Obtain the full SVD of $\widehat \Theta = [\widehat U, \widehat U_\perp] \, \widehat D \, [\widehat V, \widehat V_\perp]^\top$ where $\widehat U$ and $\widehat V$ contains the first r left-singular vectors and the first r right-singular vectors respectively. \vspace{0.07 cm}
\Stage \textbf{2: Almost Low Rank Generalized Linear Bandit}
\State Rotate the admissible parameter space: $\mathit{\Theta}^{\prime} \coloneqq [\widehat U, \widehat U_\perp]^\top \mathit{\Theta} [\widehat V, \widehat V_\perp]$, and transform the parameter set as:
\begin{align}
    \mathit{\Theta}_0 \coloneqq \{\vecc{\mathit{\Theta}^\prime_{1:r,1:r}},\vecc{\mathit{\Theta}^\prime_{r+1:d_1,1:r}},\vecc{\mathit{\Theta}^\prime_{1:r,r+1:d_2}},\vecc{\mathit{\Theta}^\prime_{r+1:d_1,r+1:d_2}}\}. \nonumber
\end{align}
\For{$t \geq T-T_1$}
\State Rotate the arm feature set: $\mathcal{X}^\prime_t \coloneqq [\widehat U, \widehat U_\perp]^\top \mathcal{X}_t [\widehat V, \widehat V_\perp]$.
\State Define the vectorized arm set so that the last $(d_1-r) \cdot (d_2-r)$ components are almost negligible:
\begin{align}
  \mathcal{X}_{0,t} \coloneqq \{\vecc{\mathcal{X}^\prime_{\{1:r,1:r\},t}},\vecc{\mathcal{X}^\prime_{\{r+1:d_1,1:r\},t}},\vecc{\mathcal{X}^\prime_{\{1:r,r+1:d_2\},t}},\vecc{\mathcal{X}^\prime_{\{r+1:d_1,r+1:d_2\},t}}\}. \nonumber
\end{align}
\State Invoke LowGLM-UCB (PLowGLM-UCB or LowUCB-GLM) with the arm set $ \mathcal{X}_{0,t}$, the parameter space $\mathit{\Theta}_0$, the low dimension $k = (d_1+d_2)r - r^2$ and penalization parameter $(\lambda_0,\lambda_\perp)$ for one round. Update the matrix $M_t(c_\mu)$ or $V_t$ accordingly.
\EndFor
\end{algorithmic}
\end{algorithm}

\begin{algorithm}[t]
\caption{Generalized Explore Subspace Then Subtract (G-ESTS)} \label{alg_ests_con}
\begin{algorithmic}[1]
\Input Action set $\{\mathcal{X}_t\}$, $T,T_1,\mathcal{D}$, the probability rate $\delta$, parameters for stage 2: $\lambda, \lambda_\perp$. \vspace{0.07 cm}
\Stage \textbf{1: Subspace Estimation}
\For{$t=1$ {\bfseries to} $T_1$}
\State Pull arm $X_t \in \mathcal{X}$ according to the distribution $\mathcal{D}$, observe payoff $y_t$.
\EndFor
\State Obtain $\widehat \Theta$ by solving the following equation:
\begin{align}
    \widehat \Theta = \arg \min_{\Theta \in R^{d_1 \times d_2}} {\dfrac{1}{T_1} \sum_{i = 1}^{T_1} \{b(\inp{X_i}{\Theta}) - y_i \inp{X_i}{\Theta}\}} + \lambda_{T_1} \nunorm{\Theta}.  \nonumber 
\end{align}
\State Obtain the full SVD of $\widehat \Theta = [\widehat U, \widehat U_\perp] \, \widehat D \, [\widehat V, \widehat V_\perp]^\top$ where $\widehat U$ and $\widehat V$ contains the first r left-singular vectors and the first r right-singular vectors respectively. \vspace{0.07 cm}
\Stage {2: Low Rank Generalized Linear Bandit} 
\State Rotate the admissible parameter space: $\mathit{\Theta}^{\prime} \coloneqq [\widehat U, \widehat U_\perp]^\top \mathit{\Theta} [\widehat V, \widehat V_\perp]$, and transform the parameter set as:
\begin{align}
    \mathit{\Theta}_0 \coloneqq \{\vecc{\mathit{\Theta}^\prime_{1:r,1:r}},\vecc{\mathit{\Theta}^\prime_{r+1:d_1,1:r}},\vecc{\mathit{\Theta}^\prime_{1:r,r+1:d_2}},\vecc{\mathit{\Theta}^\prime_{r+1:d_1,r+1:d_2}}\}. \nonumber
\end{align}
\For{$t \geq T-T_1$}
\State Rotate the arm feature set: $\mathcal{X}^\prime_t \coloneqq [\widehat U, \widehat U_\perp]^\top \mathcal{X}_t [\widehat V, \widehat V_\perp]$.
\State Define the vectorized arm set so that the last $(d_1-r) \cdot (d_2-r)$ components are almost negligible, and then drop the last $(d_1-r) \cdot (d_2-r)$ components:
\begin{align}
    \mathcal{X}_{0,sub,t} \coloneqq \{\vecc{\mathcal{X}^\prime_{\{1:r,1:r\},t}},\vecc{\mathcal{X}^\prime_{\{r+1:d_1,1:r\},t}},\vecc{\mathcal{X}^\prime_{\{1:r,r+1:d_2\},t}}\}. \nonumber
\end{align}
\State Invoke any modern generalized linear (contextual) bandit algorithm with the arm set $\mathcal{X}_{0,sub,t}$, the parameter space $\mathit{\Theta}_{0,sub}$, and the low dimension $k = (d_1+d_2)r - r^2$ for one round.
\EndFor
\end{algorithmic}
\end{algorithm}

\subsection{Algorithms for the Contextual Setting}\label{context}

To show algorithm G-ESTT and G-ESTS for the contextual setting, where the arm set $\mathcal{X}_t = \{X_{i,t}\}$ may vary over time $t = [T]$, we would firstly update some notations besides the ones we have defined in Section \ref{subsec:gestt}. We denote the time-dependent action set $\mathcal{X}_t$ after rotation as:
\begin{align}
    \mathcal{X}^\prime_t = [\widehat U, \widehat U_\perp]^\top \mathcal{X} [\widehat V, \widehat V_\perp],  \nonumber
\end{align} 
And we modify the notations of the vectorized arm set for G-ESTT and G-ESTS defined in Eqn. \eqref{rotate_arm}, \eqref{drop_arm} accordingly for each iteration:
\begin{align*}
    \mathcal{X}_{0,t} \coloneqq \{&\vecc{\mathcal{X}^\prime_{\{1:r,1:r\},t}},\vecc{\mathcal{X}^\prime_{\{r+1:d_1,1:r\},t}},\vecc{\mathcal{X}^\prime_{\{1:r,r+1:d_2\},t}},\vecc{\mathcal{X}^\prime_{\{r+1:d_1,r+1:d_2\},t}}\}, \\
    &\mathcal{X}_{0,sub,t} \coloneqq \{\vecc{\mathcal{X}^\prime_{\{1:r,1:r\},t}},\vecc{\mathcal{X}^\prime_{\{r+1:d_1,1:r\},t}},\vecc{\mathcal{X}^\prime_{\{1:r,r+1:d_2\},t}}\}.
\end{align*}

Details can be found in Algorithm \ref{alg_estt_con} and \ref{alg_ests_con}.



\section{Additional Experimental Details}\label{sec:app:exp}

\subsection{Parameter Setup for Simulations}
Here we present our parameter setting for algorithms involved in our experiment in Section \ref{sec::exp}.
\newline \textbf{Basic setup}: horizon $T=45000$. For the case where $d_1=d_2=12$ and $r=2$ we extend the horizon until $75000$ in figures to display the superiority of our proposed algorithms more clearly. The $480$ ($1000$) random matrices are sampled uniformly from $d_1d_2$-dimensional unit sphere. \vspace{2 mm}
\newline \textbf{LowESTR}: (same setup as in~\cite{lu2021low})
\begin{itemize*}
    \item failure rate: $\delta = 0.01$, the standard deviation: $\sigma = 0.01$ and the steps of stage 1: $T_1 = 1800$.
    \item penalization parameter in stage 1: $\lambda_{T_1} = 0.01\sqrt{\frac{1}{T_1}}$, and the gradient decent step size: $0.01$.
    \item $B=1, B_\perp = \frac{\sigma^2(d_1+d_2)^3 r}{T_1 D_{r,r}^2}, \lambda = 1, \lambda_\perp = \frac{T_2}{k\log(1+T_2/\lambda)}$, grid search for $\sqrt{\beta_t}$ with multiplier in $\{0.2,1,5\}$.
\end{itemize*}
\textbf{SGD-TS}: (details in~\cite{ding2021efficient})
\begin{itemize*}
    \item grid search for exploration rates in $\{0.1,1,10\}$.
    \item grid search for $C$ in $\{1,3,5,7\}$.
    \item grid search for initial step sizes in $\{0.01,0.1,1,5,10\}$.
\end{itemize*}
\textbf{G-ESTT}: (LowGLM-UCB in Stage 2)
\begin{itemize*}
    \item failure rate: $\delta = 0.01$, and the steps of stage 1: $T_1 = 1800$.
    \item $S_0=1, \mathit{\Theta} = \{X \in \mathbb{R}^{d_1 \times d_2} : \, \norm{X}_F \leq 1\}$ for the case $r=1$, and $S_0=5, \mathit{\Theta} = \{X \in \mathbb{R}^{d_1 \times d_2} : \, \norm{X}_F \leq 5\}$ for the case $r=2$.
    \item penalization in solving Eqn. \eqref{loss} with $\lambda_{T_1}$ suggested in Theorem \ref{thm_rscbound}. (We believe that a simple grid search near this value would be better.) 
    \item $p_{ij}$ set to be centered normal distribution with standard deviation $1/d$ in Stage 1. Specifically, at each round we randomly select a matrix $X_{rand,t}$ based on this $\{p_{ij}\}$ elementwisely, and then pull the arm that is closest to $X_{rand,t}$ w.r.t. $\norm{\cdot}_F$ among all candidates in the arm set.
    \item proximal gradient descent with backtracking line search solving Eqn. \eqref{loss}, step size set to $0.1$.
    \item $\lambda_0 = 1, \lambda_\perp = \frac{c_\mu^2 S_0^2 T_2}{k \log\left(1 + \frac{c_\mu S_0^2 T_2}{k \lambda_0} \right)}, S_\perp = \frac{d_1d_2r}{T_1 D_{rr}^2}\log\left(\frac{d_1+d_2}{\delta}\right)$, grid search for exploration bonus with multiplier in $\{0.2,1,5\}$.
\end{itemize*}
\textbf{G-ESTS}: (SGD-TS in Stage 2)
\begin{itemize*}
    \item The steps of stage 1: $T_1 = 1800$.
    \item penalization in solving Eqn. \eqref{loss} with $\lambda_{T_1}$ suggested in Theorem \ref{thm_rscbound}. (We believe that a simple grid search near this value would be better.) 
    \item $p_{ij}$ set to be centered normal distribution with standard deviation $1/d$ in Stage 1. Specifically, at each round we randomly select a matrix $X_{rand,t}$ based on this $\{p_{ij}\}$ elementwisely, and then pull the arm that is closest to $X_{rand,t}$ w.r.t. $\norm{\cdot}_F$ among all candidates in the arm set.
    \item proximal gradient descent with backtracking line search solving Eqn. \eqref{loss}, step size set to $0.1$.
    \item use the same setup for SGD-TS as we have listed.
\end{itemize*}

\subsection{Additonal experimental results}
\begin{figure}[t]
\begin{minipage}[b]{0.49\linewidth}
    \centering
    \includegraphics[width = 0.95\textwidth]{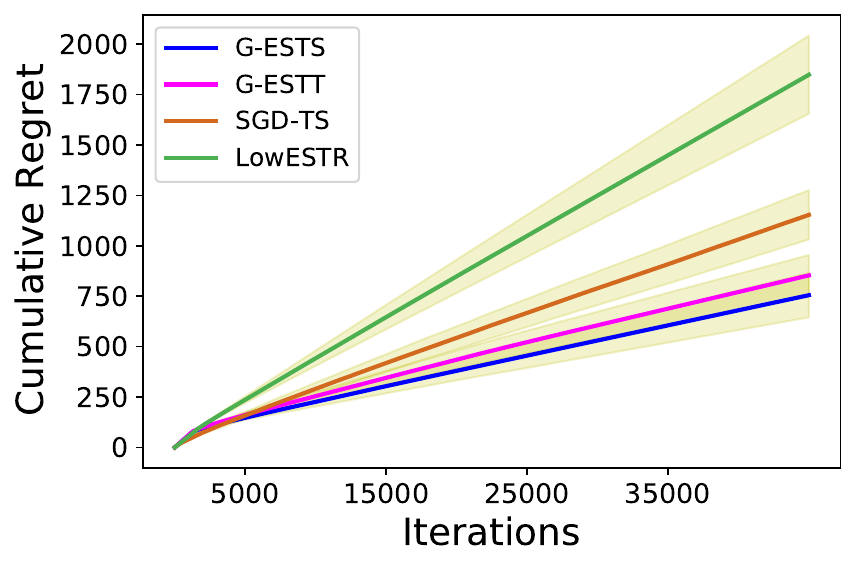}
    \vspace{-4mm}
    \captionof*{figure}{(a)}
\end{minipage}
\begin{minipage}[b]{0.49\linewidth}
    \centering
    \includegraphics[width = 0.95\textwidth]{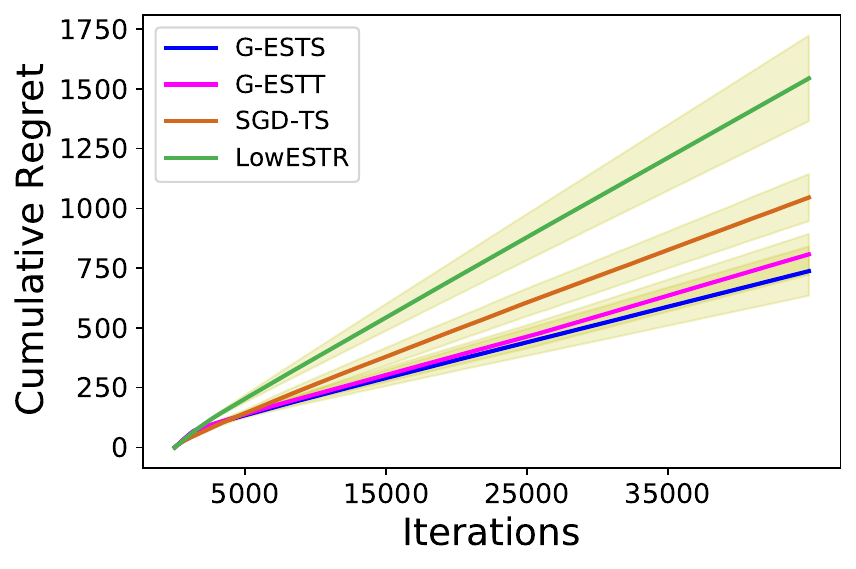}
    \vspace{-4mm}
    \captionof*{figure}{(b)}
\end{minipage}
\begin{minipage}[b]{0.49\linewidth}
    \centering
    \includegraphics[width = 0.95\textwidth]{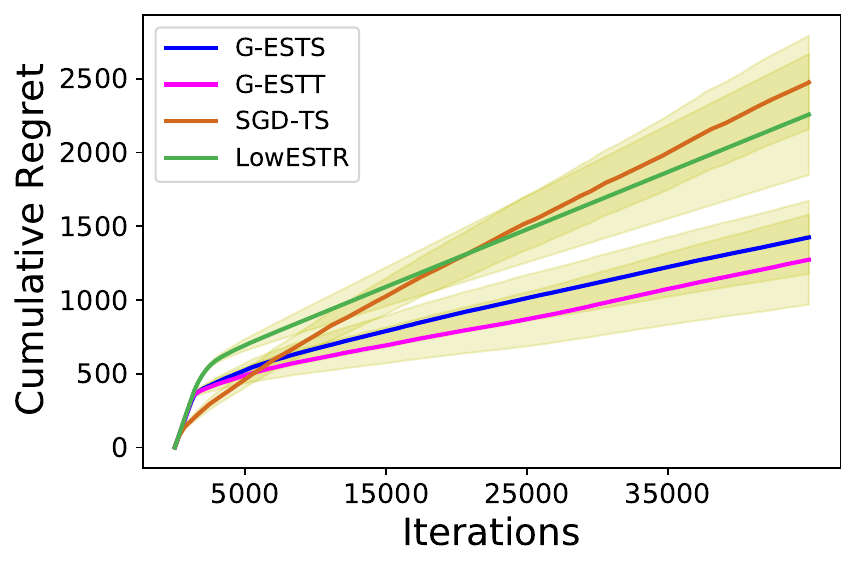}
    \vspace{-4mm}
    \captionof*{figure}{(c)}
\end{minipage}
\begin{minipage}[b]{0.49\linewidth}
    \centering
    \includegraphics[width = 0.95\textwidth]{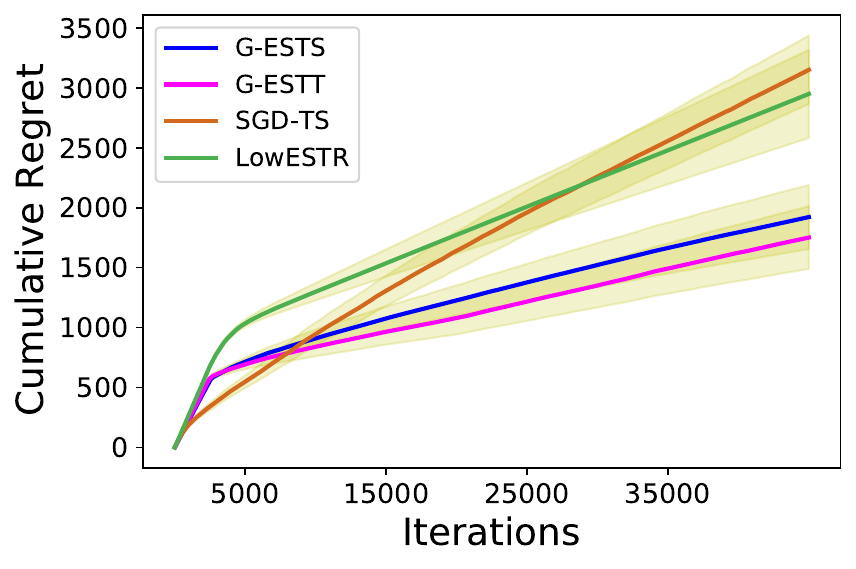}
    \vspace{-4mm}
    \captionof*{figure}{(d)}
\end{minipage}
\vspace{-3mm}
\caption{Plots of regret curves of algorithm G-ESTT, G-ESTS, SGD-TS and LowESTR under four settings ($1000$ arms). (a): diagonal $\Theta^*$ $d_1=d_2=10,r=1$; (b): diagonal $\Theta^*$ $d_1=d_2=12,r=1$; (c): non-diagonal $\Theta^*$ $d_1=d_2=10,r=2$; (d): non-diagonal $\Theta^*$ $d_1=d_2=12,r=2$.}
\label{exp2}
\end{figure}

Here we display the regret curves of algorithms under four settings with $1000$ arms in Figure \ref{exp2}, where our proposed G-ESTS and G-ESTT also dominate other methods regarding both accuracy and computation.

\subsection{Comparison between G-ESTT and G-ESTS}

In this section we compare the performance of our two frameworks G-ESTT and G-ESTS, and it is obvious that both these two proposed methods work better than the existing LowESTR and state-of-the-art generalized linear bandit algorithms under our problem setting based on Figure \ref{exp} and \ref{exp2}. Notice that G-ESTT and G-ESTS perform similarly well under the scenario $r=1$ (G-ESTS is slightly better). However, for the case $r=2$, we find that G-ESTT achieve less cumulative regret than G-ESTS does. We believe it is because that, on the one hand, G-ESTS depends more on the precision of estimate $\widehat \Theta$, which becomes more challenging for the case $r=2$. On the other hand, for G-ESTS how to reuse the random-selected actions in stage 1 is also tricky, and we will leave it as a future work. Therefore, G-ESTT (with LowUCB-GLM) quickly takes the lead in the very beginning of stage 2 since LowUCB-GLM can yield a consistent estimator early in stage 2 by reclaiming the randomly-chosen actions. 
\newline However, we find that G-ESTS is incredibly faster than other methods (including G-ESTT) as it only spends about one tenth of the running time of LowESTR until convergence as shown in Table \ref{table:time1}. Notice that G-ESTT with LowUCB-GLM is a little bit slower since it utilizes more samples for estimation in each iteration for better performance. Moreover, we conduct another simulation for the case $r=2, d_1=d_2=12$ where we additionally choose $T_1=3200$, and the results are displayed in Figure 3\footnote{Due to the size limit on arxiv, we are unable to include Figure 3 here. Please refer to the NeurIPS 2022 version.} after $100$ times repeated simulations. We observe that by appropriately enlarging the length of stage 1 $(T_1)$, G-ESTS would perform better in the long run as we expect, since a more accurate estimation of $\Theta^*$ could be obtained. Therefore, we can conclude our proposed G-ESTS could perform prominently with parsimonious computation by mildly tuning the length of stage 1 ($T_1$).  


\subsection{Comparison with other matrix subspace detection methods}\label{app:compare_lowrank}

To pre-check the efficiency of our Stein's lemma-based method for subspace estimation, we also tried the nuclear-norm regularized log-likelihood maximization with its details introduced in the following Appendix \ref{sec:app:exp}. Particularly, we could solve the regularized negative log-likelihood minimization problem with nuclear norm penalty as shown in Eqn. \eqref{loss1}. 

Specifically, we consider the two cases of our simulations: 480 arms, $d=$10, $r=$1 (Figure \ref{exp}(a) case) and 480 arms, $d=$10, $r=$2 (Figure \ref{exp}(c) case). We used the same setting as described in Appendix \ref{sec:app:exp} above ($T_1$=1800, $T$=45000), and implemented proximal gradient descent with the backtracking line search for optimization. The average regret cumulative regret along with the average transformed error $\norm{\theta^*_{(k+1):p}}_2$ defined in Eqn. \eqref{Sperp} are reported in Table \ref{table:compare_low}.

\begin{table}
\caption{Comparison between our proposed Stein's lemma-based method and the log-likelihood maximization method for low-rank matrix subspace estimation.}
\label{table:compare_low}
\vskip 0.06in
\begin{center}
\begin{small}
\begin{tabular}{lccc}
\toprule
Case & Low-rank detection method & Regret & Transformed error \\
\midrule
\multirow{2}{*}{Figure \ref{exp}(a)} &\cellcolor{gray!30} Stein's lemma-based method &\cellcolor{gray!30} G-ESTT:723.27, G-ESTS:510.80 &\cellcolor{gray!30} 0.086 \\
& Log-likelihood maximization & G-ESTT:724.96, G-ESTS:515.25 & 0.089 \\
\multirow{2}{*}{Figure \ref{exp}(c)} &\cellcolor{gray!30} Stein's lemma-based method &\cellcolor{gray!30} G-ESTT:1088.26, G-ESTS:1106.71 &\cellcolor{gray!30} 0.542 \\
& Log-likelihood maximization & G-ESTT:1136.54, G-ESTS:1198.39	 & 0.583 \\
\bottomrule
\end{tabular}
\end{small}
\end{center}
\vskip -0.04in
\end{table}

Therefore, we can see that our low-rank matrix detection method outperforms the regularized log-likelihood maximization method, especially when the underlying parameter matrix is complicated (Figure \ref{exp}(c) case). This is also consistent with our theoretical analysis, as we will show in the following Appendix \ref{sec:app:rsc} that the theoretical bound of loss $\norm{\hat\Theta - \Theta^*}^2_{\text{F}}$ is of order $d^3r/T_1$ using the regularized log-likelihood maximization method, which is worse than the convergence rate of our proposed method in Theorem \ref{thm_rscbound}.
\section{Bonus: Matrix Estimation with Restricted Strong Convexity}\label{sec:app:rsc}
\subsection{Methodology}
As we have mentioned in our main paper, we can achieve a decent matrix recovery rate regarding the Frobenius norm by using generalized first-order Stein's Lemma on Eqn. \eqref{loss}. For the completeness of our work, we also approach the matrix estimation problem by using the restricted strong convexity theory alternatively to see whether we could get the same covergence rate $O(\sqrt{{d_1+d_2)^3r}/{T_1}})$ in GLM as in the linear case under the stronger assumptions of sub-Gaussian property. Specifically, we use the regularized negative log-likelihood minimization with nuclear norm penalty for the loss function in stage 1, and consequently we are able to get the same bound as in the linear case. Notice that this work is also non-trivial since constructing the restricted strong convexity for the generalized linear low-rank matrix estimation requires us to use a truncation argument and a peeling technique~\citep{raskutti2010restricted}, which is completely different that used in simple linear case~\citep{lu2021low}. Therefore, to facilitate further study in this area and for the completeness of our work, we would present the detailed proof here in the following as a bonus. 
Loss function: we consider the following well-defined regularized negative log-likelihood minimization problem with nuclear norm penalty in stage 1:
\begin{align}
    \widehat \Theta =& \arg \min_{\Theta \in R^{d_1 \times d_2}} L_{T_1}(\Theta) 
    + \lambda_{T_1} \nunorm{\Theta}, \; \; \text{where } \nonumber \\  &L_{T_1}(\Theta) = {\dfrac{1}{T_1} \sum_{i = 1}^{T_1} \{b(\inp{X_i}{\Theta}) - y_i \inp{X_i}{\Theta}\}}, \label{loss1}
\end{align}
Different assumptions with notations reloaded: 
\begin{assumption} \label{assu_sampling1}
There exists a sampling distribution $\mathcal{D}$ over $\mathcal{X}$ with covariance matrix of $\vecc{X}$ as $\Sigma \in \R^{d_1d_2 \times d_1d_2}$, such that $\lambda_{\min}(\Sigma) = \lambda_1$ and $\vecc{X}$ is sub-Gaussian with parameter $\sigma = \lambda_2$ such that $\lambda_1/\lambda_2^2$ can be absolutely bounded. 
\end{assumption}
\begin{assumption} \label{assu_bound1}
The norm of true parameter $\Theta^*$ and feature matrices in $\mathcal{X}$ is bounded: there exists $S \in \mathbb{R}^+$ such that for all arms $X \in \mathcal{X}$, $\norm{X}_F, \norm{\Theta^*}_F \leq S$; $\norm{X}_{\text{op}}, \norm{\Theta^*}_{\text{op}} \leq S_2$ ($S_2 \leq S$).
\end{assumption}
\begin{assumption} \label{assu_link1}
The inverse link function $\mu(\cdot)$ is continuously differentiable, Lipschitz with constant $k_\mu$. $c_\mu \geq \inf_{\Theta \in \mathit\Theta, X \in \mathcal{X}} \mu^{\prime} (\inp{X}{\Theta}) > 0$ and $c_\mu \geq \inf_{\{|x| < {(S+2)\sigma c_2}\}} \mu^{\prime} (x) > 0$ for some constant $c_2$.
\end{assumption}
Here we could safely choose $\sigma = 1 / \sqrt{d_1d_2}$~\citep{lu2021low} as default. Without loss of generality, we can assume that $c_- \sigma^2 \leq \{\lambda_1, \lambda_2^2\} \leq c_+ \sigma^2$ for some absolute constant $c_-, c_+$ for the simplicity of following theoretical analysis. Assumption \ref{assu_sampling1} implies that if $X$ is sampled from the distribution $\mathcal{D}$, then for any $\Delta \in \R^{d_1 \times d_2}$ satisfying $\norm{\Delta}_F \leq 1$, we have:
\begin{align}
    \E[\inp{X}{\Delta}^2] &= \vecc{\Delta}^\top \Sigma \vecc{\Delta} \geq \lambda_1 \geq {c_- \, \sigma^2} \coloneqq \alpha; \label{alpha} \\
    &\E[\inp{X}{\Delta}^4] \leq 16 \lambda_2^4 \leq {16c_+^2 \, \sigma^4} \coloneqq \beta. \label{beta}
\end{align}

\subsection{Theorem}\label{app:rsc:thm}
\begin{theorem} \label{thm_rscbound1} \textup{(Bounds for GLM via another loss function in Eqn. \eqref{loss1})}  For any low-rank generalized linear model with samples $X_1\dots,X_{T_1}$ drawn from $\mathcal{X}$ according to $\mathcal{D}$ in Assumption \ref{assu_sampling1}, and Assumption \ref{assu_bound1}, \ref{assu_link1} hold. Then the optimal solution to the nuclear norm regularization problem \eqref{loss1} with $\lambda_{T_1} = \Omega(\sigma \sqrt{(d-\log(\delta)) /T_1})$ would satisfy:
\begin{align}
    &\norm{\widehat \Theta - \Theta^*}_F^2 \asymp \dfrac{d}{T_1 \, \sigma^2}r \asymp \dfrac{d^3 r}{T_1} , \label{thmeq_bound1} 
\end{align}
with probability at least $1-\delta$ given the condition $d\, r \lesssim \sigma^2 T_1$ and $(1+\sigma)^2 d \,r \lesssim T_1$ hold.
\end{theorem}
To prove this theorem, roughly speaking we firstly deduce the restricted strong convexity condition for our optimization problem with high probability, and then extend some previous results on the oracle inequality of estimation error.


\subsection{Restricted Strong Convexity}
\begin{definition} \label{def_rsc} \textup{(Restricted strong convexity (RSC),~\cite{negahban2012unified}).}
Given the cost function $L_{T_1}(\Theta)$ defined in \eqref{loss} and $X_1,\dots,X_{T_1} \in \R^{d_1\times d_2}$, the first-order Taylor-series error is defined as:
\begin{align*}
    \mathcal{E}_{T_1}(\Delta) \coloneqq L_{T_1}(\Theta^*+\Delta) - L_{T_1}(\Theta^*) - \inp{\nabla L_{T_1}(\Theta^*)}{\Delta}.
\end{align*}
For a given norm $\norm{\cdot}$ and regularizer $\Phi(\cdot)$, the cost function satisfies a \textit{restricted strong convexity} (RSC) condition with radius $R > 0$, curvature $\kappa > 0$ and tolerance $\tau^2$ if 
\begin{align*}
    \mathcal{E}_{T_1}(\Delta) \geq \dfrac{\kappa}{2} \norm{\Delta}^2_F - \tau_{T_1}^2\Phi^2(\Delta), \quad \ \text{for all } \norm{\Delta}_F \leq R.
\end{align*}
\end{definition} 

\begin{theorem} \label{thm_rsc} \textup{(RSC for GLM under distribution $\mathcal{D}$)}. Consider any low-rank generalized linear model with samples $X_1\dots,X_{T_1}$ drawn from $\mathcal{X}$ according to $\mathcal{D}$ in Assumption \ref{assu_sampling1}, and Assumption \ref{assu_bound1} and \ref{assu_link1} hold. Then there exists constants $c_3, c_4$ such that with probability $1-\delta$, we have the \textup{RSC} condition holds:
\textup{
\begin{align}
        \mathcal{E}_{T_1}(\Delta) \geq {c_3}\sigma^2 c_\mu \norm{\Delta}_F^2 - &(c_4 \sigma^2 + 2\sigma)\left(\sqrt{\frac{d_1}{T_1}}+\sqrt{\frac{d_2}{T_1}}\right) c_\mu \nunorm{\Delta}^2 \quad \ \text{for all } \norm{\Delta}_F \leq 1  \label{thm1_equ} \\
        \text{with } \kappa = {c_3}\sigma^2 c_\mu, \ \tau_{T_1}^2 = (c_4 \sigma^2 + &2\sigma)\left(\sqrt{\frac{d_1}{T_1}}+\sqrt{\frac{d_2}{T_1}}\right) c_\mu, \ R = 1, \ \norm{\cdot}=\norm{\cdot}_F \text{ and } \Phi(\cdot) = \nunorm{\cdot} \nonumber
\end{align}}
for $T_1 = O(\log(\log_2(d)/\delta))$.
\end{theorem}

\begin{remark} \label{rem_thm_rsc}
  The radius $R$ in Theorem \ref{thm_rsc} can be adapted to any finite positive constant keeping the same proof outline. And the required sample size $T_1$ only change in logarithmic power, which can be easily satisfied.
\end{remark}
\subsubsection{Proof of Theorem \ref{thm_rsc}}
To prove theorem \ref{thm_rsc}, we use a truncation argument and the peeling technique~\citep{raskutti2010restricted,wainwright2019high}:
\par Using the property of the remainder in the Taylor series, we have
\begin{align}
    \mathcal{E}_{T_1}(\Delta)  = \dfrac{1}{T_1} \sum_{i=1}^{T_1} \mu^{\prime} \left(\inp{X_i}{\Theta^*} + t \inp{X_i}{\Delta}\right) \inp{X_i}{\Delta}^2, \nonumber
\end{align}
for some $t \in [0,1]$. Based on \eqref{alpha} and \eqref{beta} we will set two truncation parameters $K_1^2 = 4\beta/\alpha$ and $K_2^2 = 4 \beta S^2/\alpha$ for further use. For any $\norm{\Delta}_F = \delta \in (0,1]$, we set $\tau = K_1\delta$ and a trunction function $\phi_\tau (v) = v^2 \cdot I_{\{|v| \leq 2 \tau\}}$. Then we have:
\begin{align}
    \mathcal{E}_{T_1}(\Delta) \geq \dfrac{1}{T_1} \sum_{i=1}^{T_1} \mu^{\prime} \left(\inp{X_i}{\Theta^*} + t \inp{X_i}{\Delta}\right) \phi_\tau (\inp{X_i}{\Delta}) I_{\{|\inp{X_i}{\Theta^*}| \leq K_2\}}. \nonumber
\end{align}

The right hand side would always be $0$ if $|\inp{X_i}{\Theta^*} + t \inp{X_i}{\Delta}| > 2K_1+K_2$, which implies the following result based on Assumption \ref{assu_link1}:
\begin{align}
    \mathcal{E}_{T_1}(\Delta) \geq c_\mu \dfrac{1}{T_1} \sum_{i=1}^{T_1} \phi_\tau (\inp{X_i}{\Delta}) I_{\{|\inp{X_i}{\Theta^*}| \leq K_2\}}. \nonumber
\end{align}

Therefore, it suffices to how that for all $\delta \in (0,1]$ and for $\norm{\Delta}_F = \delta$, we have:
\begin{align}
    \dfrac{1}{T_1} \sum_{i=1}^{T_1}\phi_{\tau(\delta)} (\inp{X_i}{\Delta}) I_{\{|\inp{X_i}{\Theta^*}| \leq K_2\}} \geq a_1 \delta^2 - a_2 \nunorm{\Delta} \delta, \label{trun1}
\end{align}
for some parameters $a_1$ and $a_2$ since the inequality $\norm{\Delta}_F \leq \nunorm{\Delta}$ always holds. Note the fact that $\phi_{\tau(\delta)} (\inp{X_i}{\Delta})= \delta^2\phi_{\tau(1)} (\inp{X_i}{\Delta/ \delta})$, then for any $\norm{\Delta}_F =\delta$ such that $\delta \in (0,1]$, we can apply bound \eqref{trun1} to the rescaled unit-norm matrix $\Delta/\delta$ to obtain:
\begin{align}
    \dfrac{1}{T_1} \sum_{i=1}^{T_1}\phi_\tau(1) (\inp{X_i}{\Delta / \delta}) I_{\{|\inp{X_i}{\Theta^*}| \leq K_2\}} \geq a_1  - a_2 \nunorm{\Delta / \delta},  \nonumber
\end{align}
which implies that it suffices to show \eqref{trun1} holds when $\delta = 1$, i.e.
\begin{align}
    \dfrac{1}{T_1} \sum_{i=1}^{T_1}\phi_\tau (\inp{X_i}{\Delta}) I_{\{|\inp{X_i}{\Theta^*}| \leq K_2\}} \geq a_1  - a_2 \nunorm{\Delta}, \quad \ \text{for all } \norm{\Delta}_F = 1. \nonumber
\end{align}

Then we can construct another truncation function $\tilde \phi_\tau (v)$ with parameter at most $2\tau = 2K_1$ as
\begin{align}
    \tilde \phi_\tau (v) = v^2 I_{\{|v| \leq \tau\}} + (v-2\tau)^2 I_{\{\tau < v \leq 2 \tau \}} + (v+2\tau)^2 I_{\{ -2 \tau \leq v < -\tau \}}. \nonumber
\end{align}

Then it suffices to show that
\begin{align}
    \dfrac{1}{T_1} \sum_{i=1}^{T_1} \tilde \phi_\tau (\inp{X_i}{\Delta}) I_{\{|\inp{X_i}{\Theta^*}| \leq K_2\}} \geq a_1  - a_2 \nunorm{\Delta}, \quad \ \text{for all } \norm{\Delta}_F = 1.  \nonumber
\end{align}

And for a given radius $r \geq 1$, define the random variable
\begin{align}
    Z_{T_1}(r) = \sup_{\substack{\norm{\Delta}_F = 1, \\ \nunorm{\Delta} \leq r}} \left\vert  \dfrac{1}{T_1} \sum_{i=1}^{T_1}\tilde \phi_\tau (\inp{X_i}{\Delta}) I_{\{|\inp{X_i}{\Theta^*}| \leq K_2\}} - \E\left( \tilde \phi_\tau (\inp{X}{\Delta}) I_{\{|\inp{X}{\Theta^*}| \leq K_2\}}\right)  \right\vert. \nonumber
\end{align}

Firstly, we can prove that
\begin{align}
    \E[\tilde \phi_\tau (\inp{X}{\Delta}) I_{\{|\inp{X}{\Theta^*}| \leq K_2\}}] \geq \frac{1}{2}\alpha, \label{3a/4}
\end{align}
by using the chosen values for $K_1$ and $K_2$ to show that 
\begin{align}
    \E[\tilde \phi_\tau (\inp{X}{\Delta})] \geq \frac{3}{4}\alpha, \quad \E[\tilde \phi_\tau (\inp{X}{\Delta}) I_{\{|\inp{X}{\Theta^*}| > K_2\}}] \leq \frac{1}{4}\alpha. \nonumber
\end{align}

Specifically, since we have
\begin{align*}
    \E[\tilde \phi_\tau (\inp{X}{\Delta})] \geq \E[ \inp{X}{\Delta}^2 I_{\{|\inp{X}{\Theta^*}| \leq \tau\}}] \geq \alpha - \E[ \inp{X}{\Delta}^2 I_{\{|\inp{X}{\Theta^*}| > \tau\}}]
\end{align*}

And we can show that the last term is at most $\alpha/4$ based on the Markov's inequality and Cauchy-Schwarz inequality:
\begin{align*}
    \E[ \inp{X}{\Delta}^2 I_{\{|\inp{X}{\Theta^*}| > \tau\}}] \leq \sqrt{\E[\inp{X}{\Delta}^4]} \sqrt{P([|\inp{X}{\Theta^*}| > \tau])} \leq \sqrt{\beta} \sqrt{\frac{\beta}{\tau^4}} \leq \frac{\alpha}{4}.
\end{align*}

And similarly we can prove that $\E[\tilde \phi_\tau (\inp{X}{\Delta}) I_{\{|\inp{X}{\Theta^*}| > K_2\}}] \leq \alpha/4$. On the other hand, by our choice $\tau = K_1$, the empirical process defining $Z_{T_1}(r)$ is based on functions bounded in absolute value by $K_1^2$. Thus, the functional Hoeffding inequality (Theorem 3.26 in~\cite{wainwright2019high}) implies that
\begin{align}
    P\left(Z_{T_1}(r) \geq \E(Z_{T_1}(r)) + \sigma r\left(\sqrt{\frac{d_1}{T_1}}+\sqrt{\frac{d_2}{T_1}}\right) + \frac{\alpha}{4}\right) \leq \nonumber \\  \exp\left( - \dfrac{n\left(\sigma r\left(\sqrt{\frac{d_1}{T_1}}+\sqrt{\frac{d_2}{T_1}}\right) + \frac{\alpha}{4}\right)^2}{4K_1^4}\right). \label{p_origin}
\end{align}

To bound the expected value term $\E(Z_{T_1}(r))$, we introduce an i.i.d sequance of Rademacher variables $\{\varepsilon_i\}_{i=1}^{T_1}$ and then use the symmetrization argument:
\begin{align}
    \E&(Z_{T_1}(r)) = \E\left[\sup_{\substack{\norm{\Delta}_F = 1, \\ \nunorm{\Delta} \leq r}} \left\vert  \dfrac{1}{T_1} \sum_{i=1}^{T_1}\tilde \phi_\tau (\inp{X_i}{\Delta}) I_{\{|\inp{X_i}{\Theta^*}| \leq K_2\}} - \E\left( \tilde \phi_\tau (\inp{X}{\Delta}) I_{\{|\inp{X}{\Theta^*}| \leq K_2\}}\right)  \right\vert\right] \nonumber \\
    &= \E\left[\sup_{\substack{\norm{\Delta}_F = 1, \\ \nunorm{\Delta} \leq r}} \left\vert  \dfrac{1}{T_1} \sum_{i=1}^{T_1}\tilde \phi_\tau (\inp{X_i}{\Delta}) I_{\{|\inp{X_i}{\Theta^*}| \leq K_2\}} - \E\left( \dfrac{1}{T_1} \sum_{i=1}^{T_1}\tilde \phi_\tau (\inp{Y_i}{\Delta}) I_{\{|\inp{Y_i}{\Theta^*}| \leq K_2\}}\right)  \right\vert\right] \nonumber \\
    &\leq \E_{X_i,Y_i} \left[\sup_{\substack{\norm{\Delta}_F = 1, \\ \nunorm{\Delta} \leq r}} \left\vert  \dfrac{1}{T_1} \sum_{i=1}^{T_1}\tilde \phi_\tau (\inp{X_i}{\Delta}) I_{\{|\inp{X_i}{\Theta^*}| \leq K_2\}} -  \dfrac{1}{T_1} \sum_{i=1}^{T_1}\tilde \phi_\tau (\inp{Y_i}{\Delta}) I_{\{|\inp{Y_i}{\Theta^*}| \leq K_2\}}  \right\vert\right]\nonumber \\
    &=  \E_{X_i,Y_i,\varepsilon_i} \left[\sup_{\substack{\norm{\Delta}_F = 1, \\ \nunorm{\Delta} \leq r}} \left\vert \ \dfrac{1}{T_1} \sum_{i=1}^{T_1} \varepsilon_i \left(\tilde \phi_\tau (\inp{X_i}{\Delta}) I_{\{|\inp{X_i}{\Theta^*}| \leq K_2\}} -  \tilde \phi_\tau (\inp{Y_i}{\Delta}) I_{\{|\inp{Y_i}{\Theta^*}| \leq K_2\}}\right)  \right\vert\right] \nonumber \\
    &\leq 2\E_{X_i,\varepsilon_i} \left[\sup_{\substack{\norm{\Delta}_F = 1, \\ \nunorm{\Delta} \leq r}} \left\vert \ \dfrac{1}{T_1} \sum_{i=1}^{T_1} \varepsilon_i \tilde \phi_\tau (\inp{X_i}{\Delta}) I_{\{|\inp{X_i}{\Theta^*}| \leq K_2\}}\right\vert\right] \nonumber \\
    &\stackrel{\textup{(i)}}{\leq} 8 K_1 \E_{X_i,\varepsilon_i} \left[\sup_{\substack{\norm{\Delta}_F = 1, \\ \nunorm{\Delta} \leq r}}
    \left\vert \ \dfrac{1}{T_1} \sum_{i=1}^{T_1} \varepsilon_i \inp{\Delta}{X_i} \right\vert \right] \stackrel{\textup{(ii)}}{\leq} 8 K_1 \, r \cdot
    \E_{X_i,\varepsilon_i} \left[ \norm{ \dfrac{1}{T_1} \sum_{i=1}^{T_1} \varepsilon_i {X_i} }_{\text{op}} \right], \label{e1}
\end{align}
where the inequality (i) comes from Rademacher contraction property and (ii) is by the duality between matrix $\norm{\cdot}_2$ and $\nunorm{\cdot}$ norms. Using the previous conclusion (Exercise 9.8 in~\cite{wainwright2019high}), we have 
\begin{align}
    \E_{X_i,\varepsilon_i} \left[ \norm{ \dfrac{1}{T_1} \sum_{i=1}^{T_1} \varepsilon_i {X_i} }_{\text{op}} \right] \leq \sigma c_5c_+ \left(\sqrt{\frac{d_1}{T_1}}+\sqrt{\frac{d_2}{T_1}}\right), \label{e2}
\end{align}
where $c_5$ is an independent absolute constant. Combine \eqref{p_origin}, \eqref{e1} and \eqref{e2}, we have 
\begin{align}
    P\left(Z_{T_1}(r) \geq  (8K_1c_5c_+ + 1)\sigma r\left(\sqrt{\frac{d_1}{T_1}}+\sqrt{\frac{d_2}{T_1}}\right) + \frac{\alpha}{4}\right) \leq \exp\left( - \dfrac{T_1\left(\sigma r\left(\sqrt{\frac{d_1}{T_1}}+\sqrt{\frac{d_2}{T_1}}\right) + \frac{\alpha}{4}\right)^2}{4K_1^4}\right).  \label{p_fixed}
\end{align}

According to \eqref{3a/4} and \eqref{p_fixed}, we prove the following conclusion for any fixed value of radium $r$:
\begin{align}
    P\left( \sup_{\substack{\norm{\Delta}_F = 1, \\ \nunorm{\Delta} \leq r}}\mathcal{E}_{T_1}(\Delta) < \frac{1}{4}\alpha c_\mu 
    - (8K_1c_5c_+ + 1)\left(\sqrt{\frac{d_1}{T_1}}+\sqrt{\frac{d_2}{T_1}}\right)\sigma c_\mu r \right) \leq \nonumber \\
    \exp\left( - \dfrac{T_1\left(\sigma r\left(\sqrt{\frac{d_1}{T_1}}+\sqrt{\frac{d_2}{T_1}}\right) + \frac{\alpha}{4}\right)^2}{4K_1^4}\right). \label{peeling}
\end{align}

Since we have $\norm{\Delta}_F = 1$, based on Cauchy-Schwarz inequality we have $1 \leq \nunorm{\Delta} \leq \sqrt{d}$. To prove the RSC we use a peeling argument to extend $r$ to all possible values. Define the event:
\begin{align}
    E \coloneqq \biggl\{\textup{There exists } \Delta \textup{ s.t. } \norm{\Delta}_F =1, \ \mathcal{E}_{T_1}(\Delta) &< \frac{1}{4}\alpha c_\mu 
    - (16K_1c_5c_+ + 2) \nonumber \\
    &\times \left(\sqrt{\frac{d_1}{T_1}}+\sqrt{\frac{d_2}{T_1}}\right) \sigma c_\mu \nunorm{\Delta}\biggr\}  \\
    V_i \coloneqq \{2^{i-1} \leq \nunorm{\Delta} < 2^i \}, \quad &i = 1,\dots, \left\lceil \frac{1}{2} \log_2(d) \right\rceil + 1. \nonumber
\end{align}

Then we can conclude that $E \subseteq \bigcup_{i=1}^{\left\lceil \frac{1}{2} \log_2(d) \right\rceil + 1} (E \cap V_i)$. And we can show the probability of each partition event $(E \cap V_i)$ can be upper bounded by \eqref{peeling}:
\begin{align}
    P(E \cap V_i) = 
    &P\left(   \sup_{\substack{\norm{\Delta}_F = 1, \\ 2^{i-1} \leq \nunorm{\Delta} < 2^i}}\mathcal{E}_{T_1}(\Delta) < \frac{1}{4}\alpha c_\mu 
    - (16K_1c_5c_+ + 2)\left(\sqrt{\frac{d_1}{T_1}}+\sqrt{\frac{d_2}{T_1}}\right) \sigma c_\mu \nunorm{\Delta}  \right) \nonumber \\
    &\leq   P\left(   \sup_{\substack{\norm{\Delta}_F = 1, \\ 2^{i-1} \leq \nunorm{\Delta} < 2^i}}\mathcal{E}_{T_1}(\Delta) < \frac{1}{4}\alpha c_\mu 
    - {(8K_1c_5c_+ + 1)} \left(\sqrt{\frac{d_1}{T_1}}+\sqrt{\frac{d_2}{T_1}}\right) \sigma c_\mu 2^i  \right) \nonumber \\ 
    &\leq \exp\left( - \dfrac{T_1\left(2^i\sigma \left(\sqrt{\frac{d_1}{T_1}}+\sqrt{\frac{d_2}{T_1}}\right) + \frac{\alpha}{4}\right)^2}{4K_1^4}\right), \nonumber
\end{align}
which implies that 
\begin{align}
    P(E) \leq \log_2(d) \exp\left( - \dfrac{T_1\left(2 \sigma\left(\sqrt{\frac{d_1}{T_1}}+\sqrt{\frac{d_2}{T_1}}\right) + \frac{\alpha}{4}\right)^2}{4K_1^4}\right). \nonumber
\end{align}

We complete our proof of Theorem \ref{thm_rsc} by noticing that the constants $c_3, c_4$ in \eqref{thm1_equ} only depend on the absolute constants $c_5, c_+$ and $c_-$ through our proof.  \hfill \qedsymbol
\subsection{Technical Lemmas}\label{lemmas_glm}
\begin{lemma} \label{lemma_rscbound} \textup{(Bound for GLM with nuclear regualarization,~\cite{negahban2012unified,wainwright2019high})}
Consider the negative log-likelihood cost function $L_{T_1}(\cdot)$ defined in \ref{loss} and observations $X_1,\dots,X_{T_1}$ satisfy a specific \textup{RSC} condtion in Definition 1, such that
\textup{
\begin{align}
    \mathcal{E}_{T_1}(\Delta) \geq \dfrac{\kappa}{2} \norm{\Delta}_F^2 - \tau_{T_1}^2 \nunorm{\Delta}^2, \quad \ \textup{for all } \norm{\Delta} \leq 1. \nonumber
\end{align}}
Then under the ``good" event: $\mathcal{G}(\lambda_{T_1}) \coloneqq \{\norm{\nabla L_{T_1}(\Theta^*)}_{\textup{op}} \leq \lambda_{T_1}/2 \}$, and the following two conditions hold:
\begin{align}
    \tau_{T_1}^2 r \leq \frac{\kappa}{128}, \quad 4.5\frac{\lambda_{T_1}^2}{\kappa^2}r \leq 1. \nonumber
\end{align}
Then any optimal solution to Eqn. \ref{loss1} satisfies the bound
\begin{align}
    \norm{\widehat \Theta - \Theta^*}_F^2 \leq 4.5\frac{\lambda_{T_1}^2}{\kappa^2}r. \label{lemma1bound}
\end{align}
\end{lemma}
\hfill \qedsymbol
\subsection{Proof of Theorem \ref{thm_rscbound1}}
According to Theorem \ref{thm_rsc}, there exists two absolute constants $c_3, c_4$ such that with probability at least $1-\delta$, we have the RSC condition holds:
\begin{align}
 \mathcal{E}_{T_1}(\Delta) \geq {c_3}\sigma^2 c_\mu \norm{\Delta}_F^2 - (c_4 \sigma^2 + 2\sigma)\left(\sqrt{\frac{d_1}{T_1}}+\sqrt{\frac{d_2}{T_1}}\right) c_\mu \nunorm{\Delta}^2 \quad \ \text{for all } \norm{\Delta}_F \leq 1. \nonumber
\end{align}

To implement Lemma 1, we would like to to figure out the value for regularization parameter $\lambda_{T_1}$ such that the event $\mathcal{G}(\lambda_{T_1})$ can hold with high probability and simultaneously the bound in \eqref{lemma1bound} can be well controlled.
The proof is by using the covering argument and Bernstein's inequality to bound the operator norm.
\newline Let $\xi_i = \inp{X_i}{\Theta^*}$, we have $\norm{\nabla L_{T_1}(\Theta^*)}_{\text{op}} = \norm{\frac{1}{n} \sum_{i=1}^{T_1} (b^{\prime}(\xi_i)-y_i)X_i}_{\text{op}}$, and for all $i \in [T_1]$
\begin{align}
    \E[(b^{\prime}(\xi_i)-y_i)X_i] = \E\left[X_i \, \E[b^{\prime}(\xi_i)-y_i \, \vert \, X_i]\right] = 0. \nonumber
\end{align}

Let $\mathcal{S}^{d_1} \, (\mathcal{S}^{d_2})$ be the $d_1 \, (d_2)$ dimensional Euclidean-norm unit sphere, and $\mathcal{N}^{d_1} \, (\mathcal{N}^{d_2})$ be the $1/4$ covering on $\mathcal{S}^{d_1} \, (\mathcal{S}^{d_2})$ and $\Xi(A) = \sup\limits_{\substack{u \in \mathcal{N}^{d_1}, \\ v \in \mathcal{N}^{d_2}}} u^{\top} A v$ for all $A \in \R^{d_1 \times d_2}$. By the proof of Lemma 1 in~\cite{fan2019generalized}, we know that
\begin{align}
    \norm{A}_{\text{op}} \leq \frac{16}{7} \Xi(A). \label{2normandcover}
\end{align}

Besides, based on the properties of Orlicz-1 norm and Orlicz-2 norm, we have:
\begin{align}
    \norm{(b^{\prime}(\xi_i)-y_i) u^{\top} X_i v}_{\psi_1} \leq \norm{(b^{\prime}(\xi_i)-y_i)}_{\psi_2} \norm{u^{\top} X_i v}_{\psi_2} \leq c_6 \, \sqrt{k_\mu} \lambda_2, \ \text{ for all } u \in S^{d_1}, \, v \in S^{d_2}.\nonumber
\end{align}

For some absolute constant $c_6$ (e.g. $c_6 = 6$). Then for any fixed $u \in S^{d_1}, \, v \in S^{d_2}$, by Berstein's inequality we have
\begin{align}
    P\left( \left\vert \frac{1}{T_1}\sum_{i=1}^{T_1} (b^{\prime}(\xi_i)-y_i) u^{\top} X_i v  \right\vert > t \right) \leq 2 \exp\left[-c_7 \min\left( \frac{T_1 t^2}{c_6^2 k_\mu \lambda_2^2} \, ,\frac{T_1t}{c_6 \sqrt{k_\mu} \lambda_2}\right) \right]. \nonumber
\end{align}

Then by the combination over all the union bounds and relation \eqref{2normandcover} we can claim that
\begin{align}
    P\left(\norm{ \frac{1}{T_1}\sum_{i=1}^{T_1} (b^{\prime}(\xi_i)-y_i) X_i}_{\text{op}} > \frac{16}{7}t \right) \leq 2 \, 7^{d_1+d_2} \exp\left[-c_7 \min\left( \frac{T_1 t^2}{c_6^2 k_\mu \lambda_2^2} \, ,\frac{T_1t}{c_6 \sqrt{k_\mu} \lambda_2}\right) \right]. \nonumber
\end{align}

Then the event $\{\norm{\nabla L_{T_1}(\Theta^*)}_2 \geq \frac{16}{7} t\}$ holds with probability $1- \delta$ if 
\begin{align} 
    t = \sqrt{k_\mu} \lambda_2 \max \left( \sqrt{\frac{c_6 (d_1+d_2)\log(7)+c_6 \log(2/\delta) }{T_1}} \, ,\frac{c_6 (d_1+d_2)\log(7)+c_6 \log(2/\delta) }{T_1} \right) \nonumber \\
    = \Omega\left( \sqrt{\frac{d_1+d_2-\log(\delta)}{T_1}} \, \sigma  \right). \nonumber
\end{align}

Since we assume $(d_1+d_2) \lesssim T_1$. By taking $\lambda_{T_1} = \frac{32}{7}t \asymp  \sqrt{\frac{d_1+d_2-\log(\delta)}{T_1}} \, \sigma$. We complete the proof of Theorem \ref{thm_rscbound1} and obtain the scale of the bound in \eqref{thmeq_bound1} after plugging the chosen values of $\kappa$ and $\lambda_{T_1}$ into \eqref{lemma1bound}. \hfill \qedsymbol
\newline Notice that the loss function here shown in Eqn. \eqref{loss1} is also convex and hence could be solved by a wide class of optimization methods (e.g. subgradient descent algorithm), and we have the convergence rate of matrix estimation as
$$\norm{\widehat\Theta - \Theta^*}_F = \widetilde O \left(\sqrt{\frac{d^3 r}{T_1}}\right).$$
%


\vfill

\end{document}